\documentclass[lettersize,journal]{IEEEtran}
\usepackage{amsmath,amsfonts}
\usepackage{algorithmic}
\usepackage{algorithm}
\usepackage{array}
\usepackage{subfig}
\usepackage{textcomp}
\usepackage{stfloats}
\usepackage{url}
\usepackage{verbatim}
\usepackage{graphicx}
\usepackage{cite}
\usepackage{hyperref}
\usepackage{epsfig}
\usepackage{amssymb}
\usepackage{multirow}
\usepackage{subfig}
\usepackage{bbding}
\usepackage{pifont}
\usepackage{wasysym}
\usepackage{float}
\usepackage[countmax]{subfloat}
\usepackage{caption}
\usepackage{bm}
\usepackage{colortbl}
\usepackage{xcolor} 
\usepackage{booktabs}

\usepackage{xspace}
\usepackage{color}
\usepackage{multirow}
\usepackage{adjustbox}
\usepackage{paralist} 
\usepackage{enumitem} 
\usepackage{booktabs} 
\usepackage{array}
\usepackage{caption} 

\graphicspath{{figures/}}

\renewcommand{\Paragraph}[1]{\vspace{2mm} \noindent \textbf{\textit{#1}}}

\definecolor{gray}{rgb}{0.35,0.35,0.35}
\definecolor{MyBlue}{rgb}{0,0.2,0.8}
\definecolor{MyRed}{rgb}{0.8,0.2,0}
\definecolor{MyGreen}{rgb}{0.0,0.4,0.1}
\definecolor{MyGray}{rgb}{0.4,0.4,0.4}

\long\def\ignorethis#1{}

\newlength\paramargin
\newlength\figmargin
\newlength\secmargin

\newcolumntype{L}[1]{>{\raggedright\let\newline\\\arraybackslash\hspace{0pt}}m{#1}}
\newcolumntype{C}[1]{>{\centering\let\newline\\\arraybackslash\hspace{0pt}}m{#1}}
\newcolumntype{R}[1]{>{\raggedleft\let\newline\\\arraybackslash\hspace{0pt}}m{#1}}

\usepackage{eso-pic}

\newcommand{\copyrightnotice}{%
  \AddToShipoutPictureFG*{%
    \AtPageLowerLeft{%
      \raisebox{8mm}{%
      }%
    }%
  }%
}

\begin{document}
%
\title{HPC: Hierarchical Point-based Latent Representation for Streaming Dynamic Gaussian Splatting Compression}
%
%
%

\author{Yangzhi Ma, Bojun Liu, Wenting Liao, Dong Liu,~\IEEEmembership{Senior Member,~IEEE} \\Zhu Li,~\IEEEmembership{Senior Member,~IEEE,} and Li Li$^\dag$,~\IEEEmembership{Senior Member,~IEEE,}

\thanks{Y. Ma, B. Liu, W. Liao, D. Liu and L. Li are with
MoE Key Laboratory of Brain-inspired Intelligent Perception and Cognition, University of Science and Technology of China, Hefei 230027, China (e-mail: 
mayz@mail.ustc.edu.cn;
liubj@mail.ustc.edu.cn; 
liaowenting@mail.ustc.edu.cn;
dongeliu@ustc.edu.cn;
lil1@ustc.edu.cn.)}
\thanks{Z. Li is with the University of Missouri, Kansas City. (e-mail: lizhu@umkc.edu)}
\thanks{$\dag$ denotes the corresponding author.}
}


%
%

\markboth{}%
{Shell \MakeLowercase{\textit{et al.}}: Bare Demo of IEEEtran.cls for IEEE Journals}
%




\copyrightnotice

\maketitle

\begin{abstract}
While dynamic Gaussian Splatting has driven significant advances in free-viewpoint video, maintaining its rendering quality with a small memory footprint for efficient streaming transmission still presents an ongoing challenge.
Existing streaming dynamic Gaussian Splatting compression methods typically leverage a latent representation to drive the neural network for predicting Gaussian residuals between frames.
Their core latent representations can be categorized into structured grid-based and unstructured point-based paradigms.
However, the former incurs significant parameter redundancy by inevitably modeling unoccupied space, while the latter suffers from limited compactness as it fails to exploit local correlations.
To relieve these limitations, we propose HPC, a novel streaming dynamic Gaussian Splatting compression framework.
It employs a hierarchical point-based latent representation that operates on a per-Gaussian basis to avoid parameter redundancy in unoccupied space.
Guided by a tailored aggregation scheme, these latent points achieve high compactness with low spatial redundancy.
To improve compression efficiency, \textcolor{black}{
we investigate the compression of frame-specific neural network parameters for online-optimized streaming dynamic Gaussian Splatting.
}
Combined with latent compression, this forms a fully end-to-end compression framework.
Comprehensive experimental evaluations demonstrate that HPC substantially outperforms state-of-the-art methods. It achieves a storage reduction of 67\% against its baseline while maintaining high reconstruction fidelity.
\noindent{Code is available at: \url{https://github.com/USTC-iVC-3D-Group/HPC}}

\end{abstract}

\begin{IEEEkeywords}
Dynamic Gaussian Splatting Compression, Free-viewpoint Video, Latent Representation
\end{IEEEkeywords}

%
\IEEEpeerreviewmaketitle

\section{Introduction}
\IEEEPARstart{W}{ith} the advent of immersive media, Free-Viewpoint Video (FVV) technology, which provides real-time viewing from arbitrary perspectives, has played a pivotal role in driving the evolution of next-generation media systems.
An efficient, high-quality FVV framework has the potential to unlock a variety of applications, such as virtual reality (VR) and augmented reality (AR).
For this purpose, prior works have investigated various scene representations, which involve meshes, point clouds, and radiance fields.

Among these candidates, the recently emerged 3D Gaussian Splatting (3DGS)~\cite{3dgs} is opening up substantial new avenues for FVV application.
This representation excels at handling complex visual content while supporting real-time rendering.
\textcolor{black}{
To support efficient streaming of dynamic FVV, pioneering works have extended 3DGS into an online-optimized dynamic framework~\cite{sun20243dgstream,hicom2024,fu2025recongs,chen2025motion}. 
These frameworks incrementally optimize the inter-frame deformation on a per-frame basis, enabling compact frame-by-frame scene reconstruction.
}
However, despite their essential advantages, the substantial size of the dynamic Gaussian Splatting representation poses significant challenges for streaming transmission.
To this end, integrating compression into dynamic Gaussian Splatting has become a key research focus.

In the context of compression, a prevalent strategy for dynamic Gaussians employs a latent representation coupled with several neural networks for prediction~\cite{4dgc,iFVC,zheng2025dgcpro,girish2024queen}.
The latent representation and neural networks are jointly optimized, compressed, and transmitted.
Within this framework, the design of the latent representation emerges as one of the central challenges.
Drawing inspiration from its success in earlier radiance field studies, the structural latent grid~\cite{muller2022instant,fridovich2023k} has been introduced to Gaussian Splatting~\cite{sun20243dgstream,4dgc,iFVC,zheng2025dgcpro,yu2025get3dgs}.
Within this representation, the learnable latent embeddings are allocated to the vertices of the structural grid to generate a full-space latent field.
During inference, the field is queried at the positions of individual Gaussians to retrieve their corresponding latent embeddings.
However, assigning latent embeddings across the entire space incurs a prohibitive parameter explosion, rendering it impractical for direct optimization and storage.
Therefore, such grid-based methods typically resort to either low-resolution grids~\cite{4dgc} or hashing with collisions~\cite{sun20243dgstream,iFVC,zheng2025dgcpro} to reduce parameters, which inevitably compromises the performance.

Alternatively, given that the positions of Gaussians constitute discrete sets occupying a vanishingly small fraction of the 3D space, it is \textcolor{black}{natural} to attach a latent embedding directly to each point's position, ignoring unoccupied areas. 
However, the existing point-based latent representation~\cite{girish2024queen} primarily focuses on the isolated point, failing to model correlations within local neighborhoods and thus incurring spatial redundancy.
Furthermore, given the non-uniform distribution of Gaussians, its single-scale design fails to capture the varied spatial characteristics, resulting in suboptimal performance.

In addition to the challenges in latent representation, another critical issue lies in the deficient compression strategy.
Existing studies~\cite{iFVC,4dgc,zheng2025dgcpro} have devoted to compressing the latent representation through rate-aware optimization and entropy coding, but overlook the compression of the neural network parameters.
This limitation not only prevents further bitrate reduction but also results in an imbalanced rate allocation between the latent and the network parameters.
This creates a demand for an integrated framework that co-optimizes and compresses the latent representation alongside the neural network parameters.

In response to these limitations, we propose HPC, a novel streaming dynamic Gaussian Splatting framework that integrates a \textbf{H}ierarchical \textbf{P}oint-based latent representation with a fully end-to-end \textbf{C}ompression strategy.
\textbf{On one hand}, our latent representation is built upon discrete points that operates on a per-Gaussian basis to avoid parameter redundancy in unoccupied space.
To align with the non-uniform distribution of Gaussians, we equip the representation with multi-scale receptive fields via progressive downsampling, yielding the latent hierarchy.
We further devise an aggregation scheme that incorporates both inner-scale and cross-scale processing to reduce spatial redundancy. 
It promotes local information sharing at multiple resolutions, thus achieving compactness.
\textbf{On the other hand}, \textcolor{black}{
we investigate the compression of frame-specific neural network parameters for online-optimized streaming dynamic Gaussian Splatting.
}
Through an analysis of parameter distributions, we identify potential temporal redundancy between adjacent frames.
Building on this observation, we derive a temporal reference strategy that leverages the latest decoded parameters to guide the optimization and compression of current-frame parameters.
This strategy enables us to encode the energy-compact inter-frame residual rather than full parameters, significantly lowering the bitrate consumption.
Unified with the compression of latent representation, our framework conducts a fully end-to-end optimization that balances the bitrate between the latent and network parameters, thereby attaining superior rate-distortion performance.

In summary, the proposed HPC offers the following key contributions:
\begin{itemize}
    \item The HPC framework introduces a novel hierarchical latent point representation to match the inhomogeneous distribution of Gaussians while maintaining parameter efficiency.
    Spatial redundancy is effectively reduced via a dedicated aggregation mechanism that operates both within and across scales, promoting local information sharing and resulting in a compact representation.
    \item \textcolor{black}{HPC introduces the compression of frame-specific network parameters in online-optimized streaming dynamic Gaussian Splatting.}
    The resulting temporal reference strategy encodes compact residuals and is combined with latent compression in our fully end-to-end framework, enabling optimal bitrate allocation.
    \item Extensive experimental results demonstrate that HPC achieves a remarkable rate-distortion performance against state-of-the-art methods.
    Comprehensive ablations and analyses substantiate the effectiveness of each proposed component.
\end{itemize}

The remainder of this paper is structured as follows. Section~\ref{sec:relw} reviews related work in the relevant fields. The preliminaries of our framework are provided in Section~\ref{sec:prem}. Sections~\ref{sec:representation} and~\ref{sec:compression} detail the proposed HPC methodology. Experimental configurations and results are presented and analyzed in Section~\ref{sec:exp}. Finally, Section~\ref{sec:conclusion} concludes the paper.


\section{Related Work}
\label{sec:relw}

\subsection{Static Gaussian Splatting Compression}
To address the substantial data footprint of 3DGS, Gaussian Splatting Compression seeks to eliminate redundancy from the Gaussian representations while preserving rendering fidelity.
This section presents an overview of static Gaussian Splatting compression, covering two primary branches: post-training compression and generative compression.

\textcolor{black}{Post-training compression decouples the compression process from the 3DGS training~\cite{lee2025compression,li2026gscodec,yang2025hybridgs}. By directly operating on a pre-trained 3DGS model, it enables a fast compression pipeline without introducing additional optimization overhead.}
For the optimal rate-distortion trade-off, generative compression methods perform joint training of compression and 3DGS representation~\cite{Ali2024trimming,fan2024lightgaussian,liu2024compgs,scaffoldgs,hac,contextgs,cat3dgs,chen2025hac++,li2026gscodec}.
Some methods directly reduce the number of Gaussians by employing pruning~\cite{Ali2024trimming,fan2024lightgaussian} techniques to discard sunimportant elements. 
By leveraging the local spatial redundancy of 3DGS, Scaffold-GS~\cite{scaffoldgs} constructs a more efficient representation where a region of Gaussians is clustered to a few anchor points with an implicit representation, which drastically cuts down the storage count.
Building upon Scaffold-GS, several works~\cite{hac,contextgs,cat3dgs,chen2025hac++} take one more step to introduce dedicated entropy models and end-to-end rate-distortion optimization to achieve optimal performance.


\subsection{Dynamic Gaussian Splatting Representation}
Dynamic Gaussian Splatting extends 3DGS into the temporal domain to meet the demands of FVV applications~\cite{luiten2024dynamic}.
To accommodate diverse application requirements, mainstream representations can be broadly categorized into two paradigms: offline optimization and online optimization.

\textcolor{black}{Offline methods exploit multi-frame temporal context by jointly modeling or optimizing multiple frames, enabling richer temporal dependencies across the sequence~\cite{li2024spacetime,zhang2024mega,li2025gifstream,dai20254dgv,guo2025motion}.}
In contrast, online methods process 4D scenes in an iterative manner, adapting to changes on a frame-by-frame basis to support streaming.
Representations for online dynamic Gaussians fall into two main paradigms: the structured grid and the unstructured point representation.
Following the former paradigms, 3DGStream~\cite{sun20243dgstream} utilizes a hash-grid~\cite{muller2022instant} to implicitly model inter-frame motion. Similarly, 4DGC~\cite{4dgc} employs a latent grid for motion prediction, supplementing it with a compensation strategy to reduce information loss. 
iFVC~\cite{iFVC} builds upon Scaffold-GS~\cite{scaffoldgs} and proposes a binary tri-plane for efficient residual prediction.
However, since assigning features across the entire space incurs a prohibitive memory footprint, such representations are inherently limited by hash collisions~\cite{sun20243dgstream,iFVC} or coarse resolution~\cite{4dgc}, resulting in suboptimal expressiveness.

In the point representation paradigm, parameters are directly attached to each Gaussian's position, thereby eliminating the parameter waste associated with unoccupied regions.
Following this principle, HiCoM~\cite{hicom2024} optimizes the hierarchical coherent motion for efficient prediction.
As the extension of HiCoM, ReCon-GS~\cite{fu2025recongs} introduces a dynamic hierarchy reconfiguration strategy for enhancement.
ComGS~\cite{chen2025motion} leverages a key-point motion model alongside a key-frame strategy, enabling efficient dynamic reconstruction.
Despite their advantage of rapid training, these methods suffer from limited representational compactness, as they optimize and operate on the motion directly in the raw parameter space.
Instead, another pathway is optimizing a latent representation, which relies on a neural network to learn the prediction of the motion or residual.
Within this paradigm, QUEEN~\cite{girish2024queen} operates by assigning a latent to each Gaussian and employs a jointly optimized latent decoder for residual inference. 
However, its per-point modeling scheme fails to capture correlations within local neighborhoods, incurring spatial redundancy. 
Moreover, its single-scale design is ill-suited to the non-uniformly distributed Gaussians, unable to adaptively capture spatial characteristics.

\subsection{Compression of Dynamic Gaussian Splatting and Implicit Neural Video Representation}
\textcolor{black}{
Earlier streaming NeRF methods~\cite{wang2023rerf,wu2024tetrirf} have explored temporal redundancy for compact free-viewpoint video transmission.
With the subsequent development of dynamic Gaussian representations~\cite{wang2025p4dgs,wang2024v3,li2025gifstream,dai20254dgv,girish2024queen,ke2026streamstgs}, compression methods can similarly be categorized into offline and online schemes.
}
In offline methods, MEGA~\cite{zhang2024mega} achieves compression by decoupling color attributes and employing entropy-constrained deformation.
\textcolor{black}{P-4DGS~\cite{wang2025p4dgs} exploits spatial-temporal prediction for compact 4DGS}.
Both GIFstream~\cite{li2025gifstream} and 4DGV~\cite{dai20254dgv} leverage established video codecs to achieve efficient compression.

Regarding online compression methods, QUEEN~\cite{girish2024queen} quantizes the latent representation and then sparsifies the positional residuals via a learned gating module for storage saving.
iFVC~\cite{iFVC} quantizes its tri-plane latent embeddings in binary and leverages the empirical frequency distribution for entropy estimation, thereby facilitating end-to-end rate-distortion optimization.
4DGC~\cite{4dgc} adopts a factorized entropy model~\cite{ballé2018variational} for both its motion grid and its compensated Gaussians.
Building upon this, 4DGCPro~\cite{zheng2025dgcpro} specifically enables progressive streaming and real-time coding capabilities.
\textcolor{black}{
More recently, D-FCGS~\cite{zhang2026dfcgs} adopts a feedforward GoF-based I/P coding framework with spatial-temporal entropy modeling, while StreamSTGS~\cite{ke2026streamstgs} encodes canonical Gaussian attributes and temporal features using image/video codecs for efficient streaming.
}
\textcolor{black}{
Existing online optimization-based methods primarily focus on compressing the latent representation, leaving frame-specific network parameters largely unexploited for compression.
} This omission not only hinders further bitrate reduction but also leads to an inefficient bit allocation.

To bridge this gap, relevant insights can be adopted from the field of Implicit Neural Video Representation~\cite{chen2021nerv,zhang2024boosting,wang2025uar}, where compressing the neural network itself has been a central research focus and offers well-established strategies.
\textcolor{black}{
NeRV~\cite{chen2021nerv} applies conventional model-compression techniques, including pruning, quantization, and entropy coding, directly to INR network parameters. 
Building upon this direction, Zhang et al.~\cite{zhang2024boosting} introduce a network-free entropy model based on parameter statistics, enabling differentiable rate estimation without introducing an additional entropy network. 
UAR-NVC~\cite{wang2025uar} further models dependencies among neural representation parameters through an autoregressive framework to improve compression efficiency. 
}
These studies provide useful insights for compressing the frame-specific neural networks in our streaming dynamic Gaussian representation.
Combining with latent representation compression, we are aiming to build a fully end-to-end rate-distortion optimization framework.

\section{Preliminaries}
\label{sec:prem}

\subsection{3D Gaussian Splatting}

3DGS~\cite{3dgs} achieves real-time, free-viewpoint rendering by combining a multitude of Gaussians with differentiable splatting and tile-based rasterization.
Each Gaussian is parameterized by a set of attributes. Its spatial extent and shape are defined by a mean~${\mu} \in \mathbb{R}^3$ and a covariance matrix~${\Sigma} \in \mathbb{R}^{3\times 3}$,
\begin{equation}
    G({x}) = \exp\left(-\frac{1}{2}({x}-{\mu})^\top{\Sigma}^{-1}({x}-{\mu})\right),
\end{equation}
where ${x}\in\mathbb{R}^{3}$ is an arbitrary position in the 3D space, and ${\Sigma}$ is constructed from a scaling matrix ${S}$ and a rotation matrix ${R}$ as ${\Sigma}={R}{S}{S}^\top{R}^\top$.
Following this, the rendered pixel color~${C}$ is synthesized via $\alpha$-blending, which composites these overlapping splats as follows:
\begin{equation}
    {C} = \sum_{i\in \mathcal{N}} {{c}^i \alpha^i \prod_{j=1}^{i-1} (1-\alpha^j)},
\end{equation}
where $\mathcal{N}$ is the set of sorted Gaussians contributing to the pixel, $\alpha^i$ is the opacity of the $i$-th Gaussian after projection, and ${c}^i$ is its view-dependent color obtained from SH.

\subsection{Neural Gaussian Representation}

\begin{figure*}[t]
\centering
\includegraphics[width=0.98\linewidth]{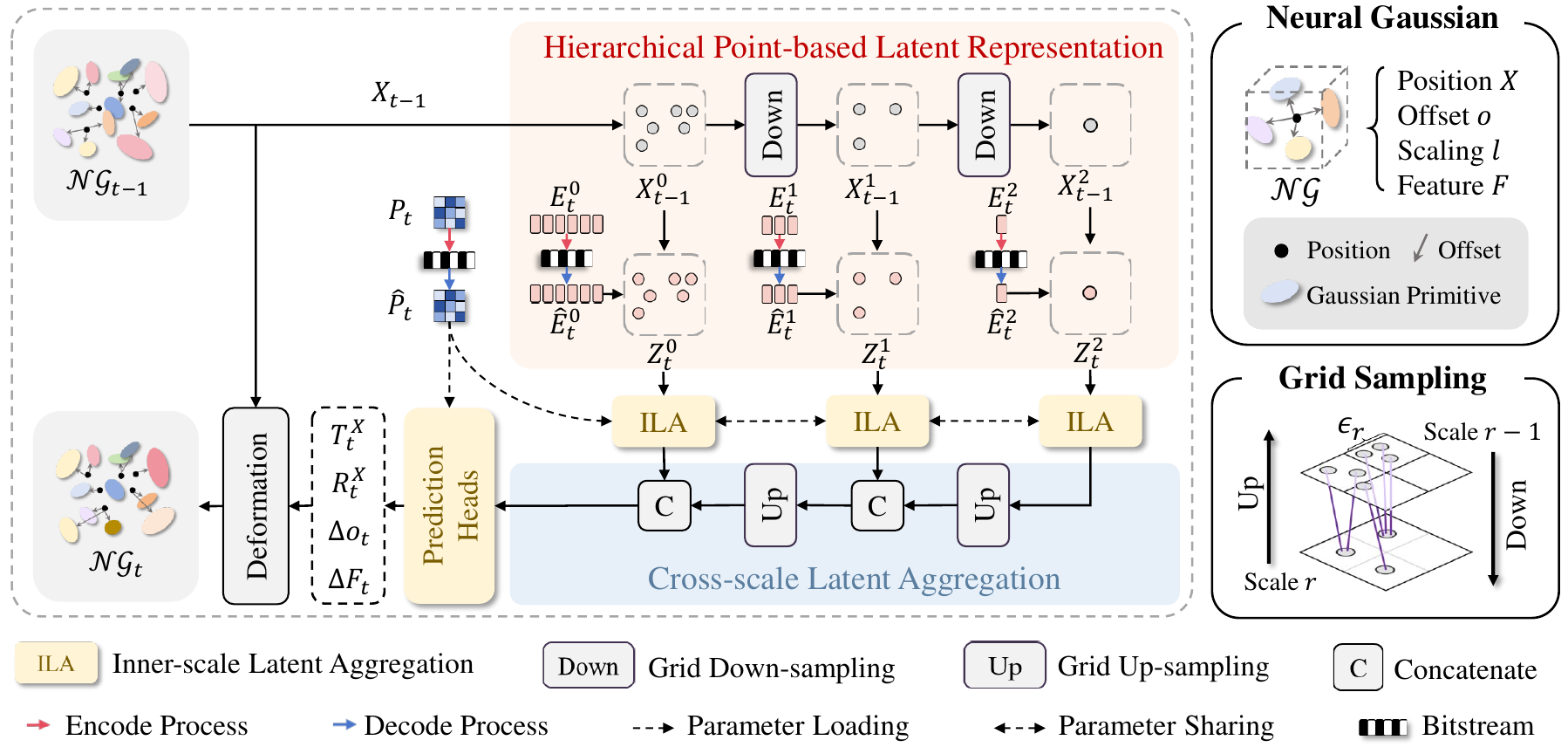}
\caption{Pipeline of the proposed HPC framework. The framework begins with the latest decoded Neural Gaussians $\mathcal{NG}_{t-1}$ as a reference. It then constructs a hierarchical latent representation $\{Z^r_t\}_{r=0}^{L-1}$ (here, $L=3$) by progressively down-sampling their positions into $\{X_{t-1}^r\}_{r=0}^{L-1}$ and pairing them with the decoded latent embeddings $\{\hat{E}_{t}^r\}_{r=0}^{L-1}$. After the Inner-scale Latent Aggregation (ILA) and Cross-scale Latent Aggregation (CLA), these latent points are fed into the prediction heads to obtain inter-frame residuals for deformation. In HPC, both the latent embeddings $\{E_{t}^r\}_{r=0}^{L-1}$ and network parameters $P_t$ are compressed for transmission. We denote the reconstructed elements from the decoder with a hat mark.}
\label{fig:overview}
\end{figure*}

Scaffold-GS~\cite{scaffoldgs} enhances the compactness and fidelity of 3DGS by eliminating its local spatial redundancy through an implicit representation. It introduces Neural Gaussians $\mathcal{NG}$ as the fundamental representation units that leverage a voxel-based clustering mechanism to manage the $M$ local Gaussians. 
Each Neural Gaussian is defined by a set of attributes:
\begin{equation}
    \mathcal{NG}=\{X \in \mathbb{R}^{3}, o \in \mathbb{R}^{3 \times M}, F \in \mathbb{R}^{D}, l\in \mathbb{R}^{3}\},
\end{equation}
where $X$ denotes the location of the anchor obtained through voxelization, $o$ represents the offsets of the $M$ managed local Gaussians, and $l$ is a scaling factor that regularizes their spatial distribution. Leveraging these attributes, the positions of the $M$ local Gaussians can be computed as:
\begin{equation}
    \{{\mu}^i\}^{M-1}_{i=0}=X+\{o^i\}^{M-1}_{i=0}\cdot l.
\end{equation}
The remaining attributes of the local Gaussians are decoupled from the implicit feature $F$ through several dedicated MLPs $\Phi_S$, conditioned on the relative viewing information $\delta$:
\begin{equation}
    \{{c}^i,{R}^i,{S}^i,\alpha^i\}^{M-1}_{i=0}=\Phi_S(F, \delta).
\end{equation}

\section{Dynamic Gaussian Representation}
\label{sec:representation}

Building upon the success of Neural Gaussians for compressing static 3DGS scenes~\cite{hac,contextgs,cat3dgs}, we extend this representation to dynamic scenes by employing a sequence of Neural Gaussians $\{\mathcal{NG}_t\}_{t=0}^{T-1}$ as the scene representation for each frame, where $t$ denotes the timestep, and $T$ is the total frame number.
Following the previous work~\cite{iFVC}, we leverage HAC~\cite{hac} to generate compact $\mathcal{NG}_0$ for the initial frame.

The reconstruction pipeline of succeeding frames is illustrated in Fig.~\ref{fig:overview}.
Given a timestep $t$, we leverage the latest decoded Neural Gaussians $\mathcal{NG}_{t-1}$ as the starting point.
To model the temporal variations efficiently, a hierarchical point-based latent representation (Sec.~\ref{sec:hlp}) is constructed by progressively down-sampling the
positions of $\mathcal{NG}_{t-1}$ and pairing them with the learnable latent embeddings.
To achieve compactness, the latent representation undergoes a two-stage aggregation process: inner-scale aggregation followed by cross-scale aggregation.
Subsequently, the aggregated latent representation guides the deformation from $\mathcal{NG}_{t-1}$ to $\mathcal{NG}_t$ (Sec.~\ref{sec:dyn_model}), reconstructing the scene of the current timestep through a motion affine transformation and feature compensation.


\subsection{Hierarchical Point-based Latent Representation}
\label{sec:hlp}
To align with the inherently discrete and non-uniform spatial distribution of Neural Gaussians, we propose using a hierarchical point-based latent representation for inter-frame prediction.
The construction of hierarchical latent points is illustrated in Fig.~\ref{fig:overview}. 
We establish the spatial hierarchy by first initializing the finest scale from the primary positions as $X^0_{t-1} = X_{t-1}$, then recursively obtaining coarser scales through grid sampling~\cite{thomas2019kpconv}:
\begin{equation}
    X^r_{t-1} = \left\lfloor X^{r-1}_{t-1}/{\epsilon^r}+0.5 \right\rfloor \cdot \epsilon^r, \quad r = 1, 2, \ldots, L-1,
\end{equation}
where $r$, $\epsilon$, and $L$ denote the scale index, grid size, and total scale count, respectively.
For each resulting position $X^r_{t-1}$, we initialize a corresponding latent embedding $E^r_t$.
After optimization, compression, and transmission, the decoder receives the reconstructed versions $\hat{E}^r_t$, which are then combined with the corresponding $X^r_{t-1}$ to construct the complete hierarchical latent points $\{Z^r_t = (X^r_{t-1}, \hat{E}^r_t)\}_{r=0}^{L-1}$.
With this design, the latent points are \textcolor{black}{able} to capture local spatial character across different scales, enabling them to better capture the temporal dynamics of non-uniformly distributed Neural Gaussians while achieving compactness for efficient storage and transmission.

To this end, we leverage the established latent representation by progressively aggregating the latent points from the coarsest to the finest scale.
We factorize this aggregation process into two complementary components: Inner-scale Latent Aggregation (ILA) and Cross-scale Latent Aggregation (CLA).

\Paragraph{Inner-scale Latent Aggregation.} 
\begin{figure}
\centering
\includegraphics[width=0.86\linewidth]{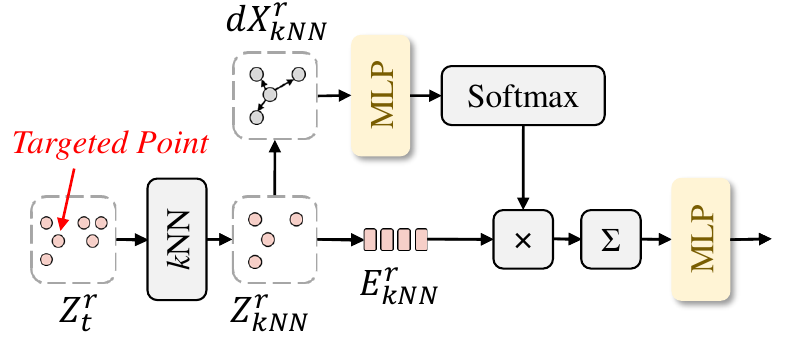}
\caption{Inner-scale Latent Aggregation (ILA). ILA takes a target point as input and locates its \textit{k}-nearest neighbors. It then predicts aggregated weights from the relative positions, performs a weighted sum of the neighbor embeddings, and finally passes the result through an MLP to produce the output.}
\label{fig:ila}
\end{figure}
Given the latent points $Z^r_t = (X^r_{t-1}, \hat{E}^r_t)$, the ILA module is designed to aggregate the embedding of a point's neighbors to model local characteristics.
As illustrated in Fig.~\ref{fig:ila}, we first introduce the $k$NN algorithm to search for the $k$ nearest neighbors for every single latent point.
\textcolor{black}{
Since the hierarchical point positions remain fixed during the optimization of each frame, the kNN neighborhoods are constructed only once per frame and reused throughout all optimization iterations.
}
Following this, the relative positional offset $dX^r_{kNN}$ between the target latent point and its neighbors is incorporated as auxiliary information to guide the aggregation.
Specifically, these relative positions are fed into an MLP to predict per-neighbor aggregation weights which are normalized via softmax.
The neighbor embeddings $\hat{E}^r_{kNN}$ are then adaptively fused according to these weights. 
Ultimately, this fusion is passed through a final MLP to integrate information across all channels, producing the aggregated result.
To reduce the parameter count, the ILA modules at different scales share the same set of parameters.
Such an aggregation scheme facilitates interaction among neighboring latent points, promoting redundancy reduction and achieving spatial compactness.

\Paragraph{Cross-scale Latent Aggregation.}
Building upon ILA, we next introduce the CLA process to fuse information across different resolutions. 
As illustrated in Fig.~\ref{fig:overview}, CLA operates the inner-scale-aggregated latent points progressively from coarse to fine scales. 
At each step, coarse-scale embeddings are upsampled by copying them to their corresponding fine-scale positions, thereby recovering resolution. 
These upsampled points are then concatenated with the native fine-scale latent points to enable cross-scale fusion. 
Such a process is repeated recursively until the original resolution is reached, ultimately producing the cross-scale aggregated latent points for subsequent prediction.
Through this CLA process, the aggregated latent representation integrates local spatial context captured across multiple receptive fields. 
Such a powerful context enables it to effectively model the underlying characteristics of non-uniformly distributed Neural Gaussians, thereby providing a superior foundation for subsequent prediction.

\subsection{Dynamic Neural Gaussian Deformation}
\label{sec:dyn_model}

Within the aggregated latent points, we proceed to predict the Neural Gaussians $\mathcal{NG}_t=\{X_t, o_t, F_t, l_t\}$ for the current timestep, conditioned on the previous state $\mathcal{NG}_{t-1}=\{X_{t-1}, o_{t-1}, F_{t-1}, l_{t-1}\}$. 
In this deformation, the scaling factor $l$ is held constant as its variation has been proven to degrade performance~\cite{iFVC}, whereas $X$, $o$, and $F$ are updated. 
Given this setup, the deformation is implemented via two complementary mechanisms: a residual compensation for the implicit feature and a motion affine transformation for geometric deformation.

Specifically, the aggregated latent points are first passed through the MLP-based prediction heads, yielding a set of deformation parameters: feature residual $\Delta F_t$, anchor translation $T^X_t$, anchor rotation $R^X_t$, and offset residual $\Delta o_t$. 
For the implicit features, the compensation is achieved through adding the predicted residual $\Delta F_t$:
\begin{equation}
    F_t = F_{t-1} + \Delta F_t.
\end{equation}

Geometric deformation is modeled by a motion affine transformation that combines global and local adjustments. 
This transformation updates the anchor position through translation $T^X_t$.
Additionally, each local offset receives an extra individual adjustment $\Delta o_t$. 
To enable a more expressive geometric transformation, we adopt the strategy from~\cite{li2025gifstream} of applying a global rotation $R^X_t$ to all corresponding offsets.
Taken together, the complete update can be concisely expressed as:
\begin{equation}
        X_t = X_{t-1} + T^X_t,
\end{equation}
\begin{equation}
        o_t = R^X_t (o_{t-1} + \Delta o_t).
\end{equation}
Under such a mechanism, the Neural Gaussians gain sufficient expressive power to model temporal variation, enabling high-fidelity rendering through Neural Gaussian Splatting~\cite{scaffoldgs}.

\section{Compression Scheme}
\label{sec:compression}

\begin{figure*}[t]
\centering
\includegraphics[width=0.9\linewidth]{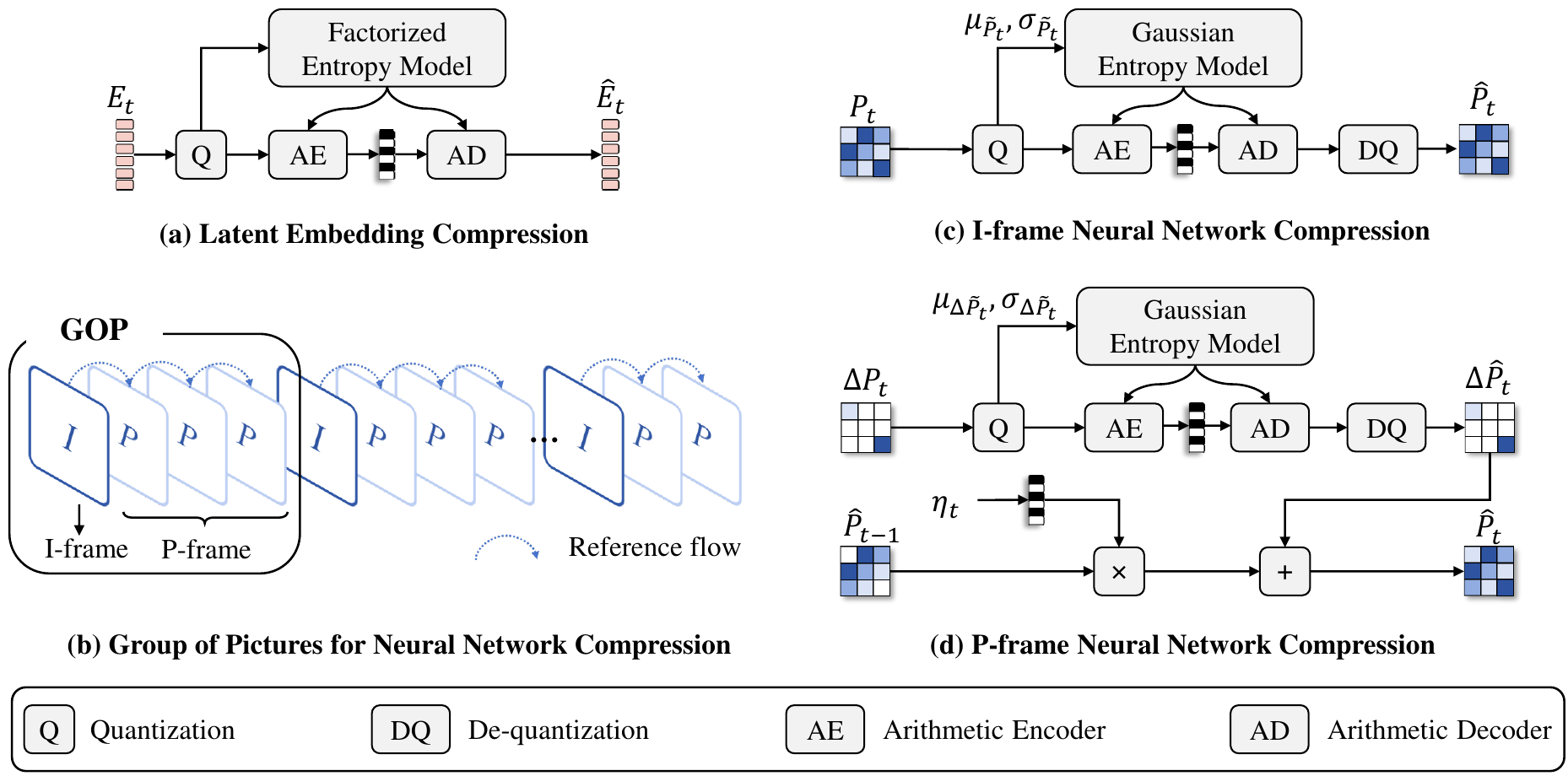}
\caption{HPC's compression scheme. We incorporate compression for both the latent embeddings and the neural networks.}
\label{fig:compression}
\end{figure*}

A tailored compression scheme is integrated into HPC, aiming to fulfill the transmission requirements of streaming FVV.
In our framework, the elements to be compressed and transmitted are the latent embedding $E_t=\{E^r_t\}_{r=0}^{L-1}$ (Sec.~\ref{sec:lec}) and the neural network parameters $P_t$  (Sec.~\ref{sec:nnc}).
Especially, the compression of neural network parameters has been largely overlooked in streaming dynamic Gaussian Splatting, where existing methods~\cite{sun20243dgstream,iFVC,4dgc,girish2024queen,zheng2025dgcpro} simply transmit full-precision 32-bit floats. To bridge this gap and enhance rate-distortion efficiency, our work presents a novel scheme for compressing the streaming neural network parameters.
Finally, we combine the compression of both parts and derive the overall rate-distortion objective to realize a fully end-to-end optimization (Sec.~\ref{sec:rdo}).

\subsection{Latent Embedding Compression}
\label{sec:lec}
By virtue of the ILA and CLA design, the optimized latent embeddings~${E}_t$ are compact with reduced spatial redundancy, ensuring they are well-suited for compression.
As illustrated in Fig.~\ref{fig:compression}~(a), the compression pipeline includes a quantization operation that discretizes the latent embeddings into integers, followed by an entropy coding module.

To circumvent the non-differentiability of quantization during training, we adopt the common practice from prior works~\cite{Video2023Gomes,NIRVANA2023Maiya,zhang2024boosting} of employing distinct quantization proxies for distortion calculation and entropy estimation to obtain the quantized embedding $\hat{E}_t$.
Specifically, the straight-through estimator (STE) is adopted for distortion calculation:
\begin{equation}
    \hat{E}_t=\text{SG}(\lfloor E_t \rceil - E_t) + E_t,
\end{equation}
where $\text{SG}(\cdot)$ denotes the gradient-stopping operation.
For entropy estimation, the rounding quantization is substituted with adding a uniform noise $u$:
\begin{equation}
    \hat{E}_t = 
    E_t + u, \quad u \sim \mathcal{U}(-\frac{1}{2}, \frac{1}{2}).
\end{equation}

After quantization, the widely used factorized model~\cite{ballé2018variational} is adopted for entropy estimation. 
The entropy model comprises several learnable layers to progressively refine the input embedding to calculate their probabilities and bitrate:
\begin{equation}
    p_{\text{PMF}}(\hat{E}_t) =
    p_{\text{CDF}}(\hat{E}_t+\frac{1}{2})-p_{\text{CDF}}({\hat{E}_t-\frac{1}{2}}),
\end{equation}
\begin{equation}
    \mathcal{R}(\hat{E}_t)=\mathbb{E}_{\hat{E}_t}[-\log_2 p_{\text{PMF}}(\hat{E}_t)],
\end{equation}
where $p_{\text{PMF}}(\cdot)$ is the probability mass function, $p_{\text{CDF}}(\cdot)$ is the cumulative distribution function approximated by the factorized model, and $\mathcal{R}(\hat{E}_t)$ is the estimated bitrate of $\hat{E}_t$.
Within the approximated probabilities, we can apply entropy coding for the latent embeddings to achieve further compression.
\textcolor{black}{
For actual coding, one global minimum/maximum pair is transmitted for each frame to recover the quantization range of the latent embeddings.
}

\subsection{Neural Network Compression}
\label{sec:nnc}
\begin{figure*}[]
\centering
\includegraphics[width=0.96\linewidth]{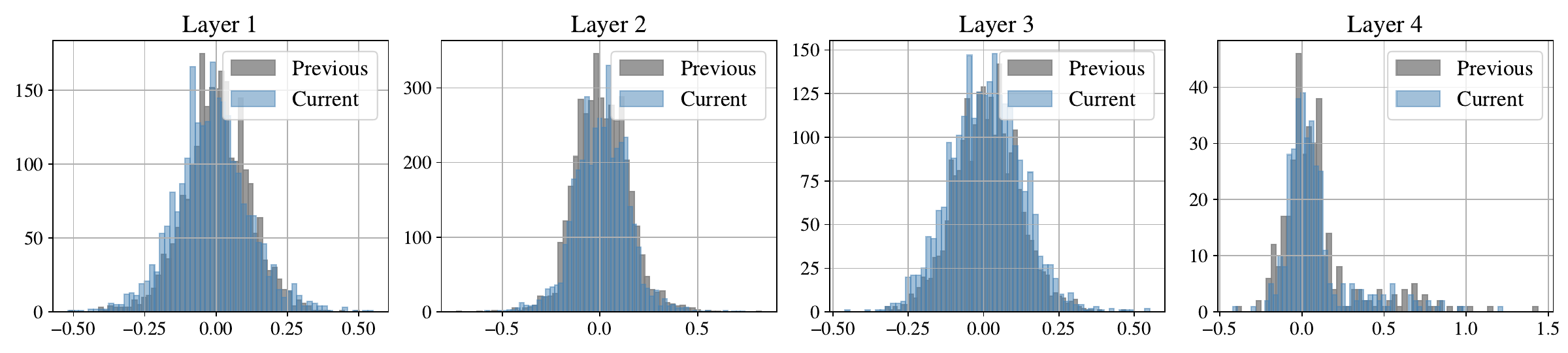}
\caption{Parameter distributions across adjacent frames in different layers.}
\label{fig:nnhist}
\end{figure*}

To explore optimal compression of the neural networks in HPC, we start with analyzing the characteristics of their parameters.
\textcolor{black}{
Since independently optimized networks do not provide explicit element-wise parameter alignment across frames, direct element-wise correlation is not meaningful at this stage. Nevertheless, the similar marginal distributions of corresponding layers in Fig.~\ref{fig:nnhist} heuristically suggest similar functional roles, motivating us to impose constraints between corresponding parameter elements across adjacent frames to induce exploitable temporal correlation.
}

However, implementing such a naive strategy may introduce error propagation, where poorly optimized parameters in one frame can adversely affect subsequent frames, resulting in cascading performance decline.
Drawing inspiration from traditional video coding, we refer to the concept of Group of Pictures (GOP)~\cite{sullivan2012overview} for an optimal reference structure.
Specifically, as illustrated in Fig.~\ref{fig:compression} (b), a GOP typically comprises an Intra-coded frame (I-frame) and several Predictive-coded frames (P-frames).
The I-frame, positioned at the GOP start, is independently optimized and compressed, while subsequent P-frames employ inter-prediction by referencing the preceding frame.
By introducing a GOP structure with periodic insertion of independent I-frames, we mitigate error propagation and maintain optimization stability over time.

\Paragraph{I-frame Neural Network Compression.}
As illustrated in Fig.~\ref{fig:compression}~(c), the compression pipeline for I-frames comprises quantization followed by entropy coding.
Since the network parameters are particularly sensitive to quantization errors, we introduce a scaling operation prior to quantization to better preserve their precision:
\begin{equation}
    \tilde{P_t} = \lfloor \frac{{P}_t-\min({P}_t)}{\max({P}_t)-\min({P}_t)} \cdot (2^{B}-1) \rceil,
\end{equation}
where $\tilde{P_t}$ denotes the quantized and transmitted parameters, and $B$ is the bit depth of quantization which controls the quantization precision.
\textcolor{black}{
In practice, quantization is performed independently for each parameter tensor, with the weight and bias tensors of each layer treated with their own quantization range.
}
The minimum and maximum values of ${P}_t$ are sent to the decoder as side information to obtain the reconstructed parameters $\hat{P}_t$:
\begin{equation}
    \hat{P}_t = {\tilde{P}_t \cdot [\max({P}_t)-\min({P}_t)]}/{(2^{B}-1)} + \min({P}_t).
\end{equation}
During training, we adopt the same proxy strategy used for latent embeddings: the STE for gradient propagation in the distortion term, and additive uniform noise for differentiable entropy estimation. 

After quantization, entropy coding is applied to $\tilde{P}_t$ for the final compression. 
Following~\cite{zhang2024boosting}, we introduce the network-free Gaussian distribution for entropy modeling, which is parameterized solely by the statistical mean $\mu_{\tilde{P}_t}$ and variance $\sigma^2_{\tilde{P}_t}$.
Consequently, the approximated probabilities $p_{\text{PMF}}(\tilde{P_t})$ and bitrate $\mathcal{R}(\tilde{P_t})$ are available as:
\begin{equation}
    p_{\text{PMF}}(\tilde{P_t}) =
    \prod_{i}(\mathcal{N}(\mu_{\tilde{P}_t}, \sigma^2_{\tilde{P}_t}) * \mathcal{U}(-\frac{1}{2},\frac{1}{2}))(\tilde{P}_t^{i}),
\end{equation}
\begin{equation}
    \mathcal{R}(\tilde{P_t})=\mathbb{E}_{\tilde{P}_t}[-\log_2 p_{\text{PMF}}(\tilde{P}_t)],
\end{equation}
where $*$ denotes convolution, and $P_t^{i}$ denotes the \textcolor{black}{$i$-th} element in $P_t$.
This design enables differentiable entropy estimation without introducing additional network parameters.
\textcolor{black}{
In actual coding, network quantization and entropy coding are performed at the parameter-tensor level, with the weight and bias tensors processed separately. Each transmitted tensor uses its own quantization range and Gaussian statistics. The above metadata is computed from the actual parameter tensor being encoded and is explicitly transmitted to the decoder.
}

\Paragraph{P-frame Neural Network Compression.}
For P-frames, we aim to exploit the decoded parameters from the previous frame ${\hat{P}}_{t-1}$ as a contextual prior, which guides the optimization and compression of the current parameters.
To this end, we draw upon the classical differential coding to optimize and encode the residual between consecutive frames, which exhibits greater energy compaction.

The P-frame neural network compression is illustrated in Fig.~\ref{fig:compression}~(d). 
Specifically, inspired by \textcolor{black}{the similar layer-wise marginal distributions observed} in Fig.~\ref{fig:nnhist}, we first introduce a learnable, per-layer scaling factor $\eta_t$ to adapt $\hat{P}_{t-1}$ for a coarse prediction of the current parameters. 
To bridge the remaining gap, we further optimize a per-parameter residual $\Delta P_t$ for fine-grained compensation.
Both $\eta_t$ and $\Delta P_t$ are transmitted to the decoder. Due to its minimal count, the per-layer $\eta_t$ is preserved at the original precision without compression. In contrast, the per-parameter $\Delta P_t$ undergoes further compression for bitrate reduction.
Within the received $\eta_t$ and the decoded residuals $\Delta \hat{P}_t$, the final parameters $\hat{P}_t$ can be reconstructed as follows:
\begin{equation}
\label{form:paramcomp}
    \hat{P}_t = \eta_t \cdot \hat{P}_{t-1} + \Delta \hat{P}_t.
\end{equation}
\textcolor{black}{
 The layer-wise scaling factors $\eta_t$ are additionally transmitted in full precision.}

To compress the residuals $\Delta P_t$, we apply a similar quantization strategy as used for the I-frame parameters. 
\textcolor{black}{
The quantization range is jointly determined by $\hat{P}_{t-1}$ and $\Delta P_t$. Since $\Delta P_t$ typically has a much smaller dynamic range, the resulting quantization step is approximately aligned with that of the full parameters, while the residual values remain concentrated within a narrow subset of quantization levels for bit rate saving:
}
\begin{equation}
    v_\text{min} = \min({\hat{P}}_{t-1}, {\Delta P_t}), \quad v_\text{max} = \max({\hat{P}}_{t-1}, {\Delta P_t}).
\end{equation}
\begin{equation}
    \Delta \tilde{P_t} = \lfloor \frac{{\Delta P_t}-v_\text{min}}{v_\text{max}-v_\text{min}} \cdot (2^{B}-1) \rceil,
\end{equation}
\begin{equation}
    \Delta \hat{P}_t = {\Delta \tilde{P}_t \cdot (v_\text{max}-v_\text{min})}/({2^{B}-1}) + v_\text{min}.
\end{equation}
\textcolor{black}{
Including $\Delta P_t$ in the joint range also avoids clipping when occasional residual values exceed the range of $\hat{P}_{t-1}$.
}For further entropy calculation, we adopt the same strategy as the compression of I-frame parameters:
\begin{equation}
    p_{\text{PMF}}(\Delta \tilde{P_t}) =
    \prod_{i}(\mathcal{N}(\mu_{\Delta \tilde{P}_t}, \sigma^2_{\Delta \tilde{P}_t}) * \mathcal{U}(-\frac{1}{2},\frac{1}{2}))(\Delta \tilde{P_t}^{i}),
\end{equation}
\begin{equation}
    \mathcal{R}(\Delta \tilde{P_t})=\mathbb{E}_{\Delta \tilde{P_t}}[-\log_2 p_{\text{PMF}}(\Delta \tilde{P_t})].
\end{equation}
\textcolor{black}{
For P-frames, the same tensor-wise coding procedure is applied.
}
Within the proposed temporal reference strategy, the optimized residuals exhibit higher energy compactness. This leads to a lower transmission cost compared to encoding the full parameters, while still preserving the representational capacity of the neural network.

\subsection{Rate-distortion Optimization}
\label{sec:rdo}
Finally, we jointly optimize the entire framework by minimizing a combined loss of the total estimated bitrate and the rendering distortion, pursuing the optimal rate-distortion performance in an end-to-end manner.
Given a timestep $t$, we calculate the weighted sum of distortion loss $\mathcal{L}^{(t)}_\mathcal{D}$ and the rate loss $\mathcal{L}^{(t)}_\mathcal{R}$ as our overall supervision objective $\mathcal{L}^{(t)}$:
\textcolor{black}{
\begin{equation}
    \mathcal{L}^{(t)} = \mathcal{L}^{(t)}_\mathcal{D}+\lambda \mathcal{L}^{(t)}_\mathcal{R},
\end{equation}
}
where $\lambda$ is the Lagrange multiplier which controls the rate-distortion balance.
Specifically, we adopt the 3DGS rendering loss~\cite{3dgs} as the distortion term:
\begin{equation}
    \mathcal{L}^{(t)}_\mathcal{D} = \mathcal{L}^{(t)}_1 + \lambda_{SSIM} \mathcal{L}^{(t)}_{SSIM}.
\end{equation}
For the rate term, we combined the estimated bitrate of both latent embeddings and neural network parameters:
\begin{equation}
    \mathcal{L}^{(t)}_\mathcal{R}=
    \begin{cases}
        \mathcal{R}(\hat{E}_t)+\mathcal{R}(\tilde{P}_t), & t \bmod T_{GOP} = 0, \\
        \mathcal{R}(\hat{E}_t)+\mathcal{R}(\Delta \tilde{P}_t), & t \bmod T_{GOP} \neq 0,
    \end{cases}
\end{equation}
where $T_{GOP}$ is the predefined size of a GOP.
\textcolor{black}{
We note that the factorized entropy-model parameters are directly quantized and entropy-coded at every frame, since applying the same parameter-rate constraint to them increases the latent bitrate and slightly degrades the overall rate-distortion performance. 
}

\section{Experiments}
\label{sec:exp}

\begin{table}[]
\centering
\caption{Quantitative comparison. 
The online methods are organized, in order, into three categories: grid-based methods, existing point-based methods, and our methods, with horizontal lines separating each category.
For methods with multiple rate-distortion points, we report both the lowest (-\textit{l}) and highest (-\textit{h}) quality results. The best and second-best results are highlighted in \colorbox{red!25}{Red} and \colorbox{orange!25}{Orange} cells, respectively.
\textcolor{black}{
Unavailable entries correspond to results not reported in the original paper.
}}
\begin{tabular}{@{}clccc@{}}
\toprule
\multicolumn{5}{c}{N3DV dataset} \\ \midrule
Category & \multicolumn{1}{c}{Method} & \begin{tabular}[c]{@{}c@{}}PSNR\\ (dB)$\uparrow$\end{tabular} & SSIM$\uparrow$ & \begin{tabular}[c]{@{}c@{}}Storage\\ (KB/frame)$\downarrow$\end{tabular} \\ \midrule
 & STG~\cite{li2024spacetime} & 32.05 & -  & 666 \\
 & GIFStream~\cite{li2025gifstream} & 31.75 & \cellcolor{orange!25}{0.938} & \cellcolor{red!25}{33} \\
 & Ex4DGS~\cite{lee2024ex4dgs} & \cellcolor{orange!25}{32.11} & \cellcolor{red!25}{\textcolor{black}{0.940}} & 383 \\
 & MEGA~\cite{zhang2024mega} & 31.49 & -  & 83 \\
\multirow{-6}{*}{\textcolor{black}{Offline}} & 4DGV~\cite{dai20254dgv} & \cellcolor{red!25}{32.55} & -  & \cellcolor{orange!25}{70} \\ \midrule
 & {3DGStream~\cite{sun20243dgstream}} & 31.69 & 0.948 & 7780 \\
 & {4DGC~\cite{4dgc}} & 31.58 & 0.943 & 500 \\
 & {4DGCPro~\cite{zheng2025dgcpro}-\textit{l}} & 30.68 & 0.926 & 210 \\
 & {4DGCPro~\cite{zheng2025dgcpro}-\textit{h}} & 31.64 & 0.944 & 640 \\
 & {iFVC~\cite{iFVC}-\textit{l}} & 31.84 & 0.950 & 63 \\
 & {iFVC~\cite{iFVC}-\textit{h}} & 32.35 & 0.953 & 99 \\ \cmidrule{2-5}
 & HiCoM~\cite{hicom2024} & 31.17 & -  & 900 \\
 & QUEEN~\cite{girish2024queen}-\textit{l} & 31.89 & 0.946 & 680 \\
 & QUEEN~\cite{girish2024queen}-\textit{h} & 32.19 & 0.946 & 750 \\
 & ComGS~\cite{chen2025motion}-\textit{l} & 31.87 & -  & 49 \\
 & ComGS~\cite{chen2025motion}-\textit{h} & 32.12 & -  & 106 \\
 & ReCon-GS~\cite{fu2025recongs} & \cellcolor{red!25}{32.66} & \cellcolor{red!25}{0.957} & 440 \\  
 & \textcolor{black}{D-FCGS}~\cite{zhang2026dfcgs} & \textcolor{black}{31.91} & \textcolor{black}{0.952} & \textcolor{black}{460} \\
 & \textcolor{black}{StreamSTGS}~\cite{ke2026streamstgs} & \textcolor{black}{32.30} & \textcolor{black}{0.943} & \textcolor{black}{173} \\\cmidrule{2-5}
 & HPC w/o NNC-\textit{l} (Ours) & 32.22 & 0.953  & {59} \\
 & HPC w/o NNC-\textit{h} (Ours) & \cellcolor{orange!25}{32.36} & \cellcolor{orange!25}{0.955}  & {66} \\
 & HPC-\textit{l} (Ours) & 31.91 & 0.951  & \cellcolor{red!25}{23} \\
\multirow{-12}{*}{Online} & HPC-\textit{h} (Ours) & \cellcolor{orange!25}{32.36} & \cellcolor{orange!25}{0.955}  & \cellcolor{orange!25}{39} \\ \midrule
\multicolumn{5}{c}{Meet Room dataset} \\ \midrule
Category & \multicolumn{1}{c}{Method} & \begin{tabular}[c]{@{}c@{}}PSNR\\ (dB)$\uparrow$\end{tabular} & SSIM$\uparrow$ & \begin{tabular}[c]{@{}c@{}}Storage\\ (KB/frame)$\downarrow$\end{tabular} \\ \midrule
 & STG~\cite{li2024spacetime} & 29.51 & 0.932  & \cellcolor{orange!25}{238} \\
 & Ex4DGS~\cite{lee2024ex4dgs} & \cellcolor{orange!25}{31.03} & \cellcolor{orange!25}{0.946} & 250 \\
\multirow{-3}{*}{Offline} & 4DGV~\cite{dai20254dgv} & \cellcolor{red!25}{32.31} & \cellcolor{red!25}{0.957} & \cellcolor{red!25}{64} \\ \midrule
 & 3DGStream~\cite{sun20243dgstream} & 30.79 & -  & 4100 \\
 & 4DGC~\cite{4dgc} & 28.08 & 0.922 & 420 \\
 & iFVC~\cite{iFVC}-\textit{l} & 32.05 & 0.956 & 63 \\
 & iFVC~\cite{iFVC}-\textit{h} & {32.39} & {0.959} & 80 \\ \cmidrule{2-5}
 & HiCoM~\cite{hicom2024} & 26.73 & -  & 600 \\
 & ComGS~\cite{chen2025motion} & 31.49 & -  & \cellcolor{orange!25}{28} \\
 & ReCon-GS~\cite{fu2025recongs} & 30.84 & 0.954 & 300 \\ 
 & \textcolor{black}{StreamSTGS}~\cite{ke2026streamstgs} & \textcolor{black}{27.41} & \textcolor{black}{0.920} & \textcolor{black}{142} \\\cmidrule{2-5}
 & HPC w/o NNC-\textit{l} (Ours) &32.26 & 0.959  & 52 \\
 & HPC w/o NNC-\textit{h} (Ours) &\cellcolor{red!25}{32.62} & \cellcolor{red!25}{0.962}  & 61 \\
 & HPC-\textit{l} (Ours) &31.71 & 0.957  & \cellcolor{red!25}{18} \\
\multirow{-10}{*}{Online} & HPC-\textit{h} (Ours) &\cellcolor{orange!25}{32.57} & \cellcolor{orange!25}{0.961}  & \cellcolor{orange!25}{28} \\ \bottomrule
\end{tabular}
\label{tab:benchmark}
\end{table}

\begin{table}[]
\centering
\caption{BD-BR (\%) results against iFVC~\cite{iFVC}. The results are tested on the N3DV and Meet Room datasets.}
\begin{tabular}{@{}lccc@{}}
\toprule
Method & N3DV & Meet Room & Average \\ \midrule
HPC & \textbf{-64.9} & \textbf{-70.7} & \textbf{-67.8} \\ 
HPC w/o NNC & -27.1 & -28.9 & -28.0 \\ \bottomrule
\end{tabular}
\label{tab:ablcomponent}
\end{table}

\begin{figure*}[t]
	\centering
	\includegraphics[width=0.49\linewidth]{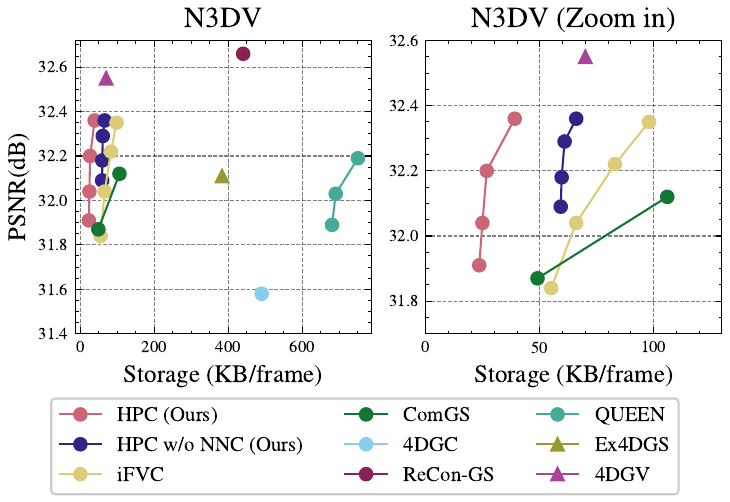}
    \includegraphics[width=0.49\linewidth]{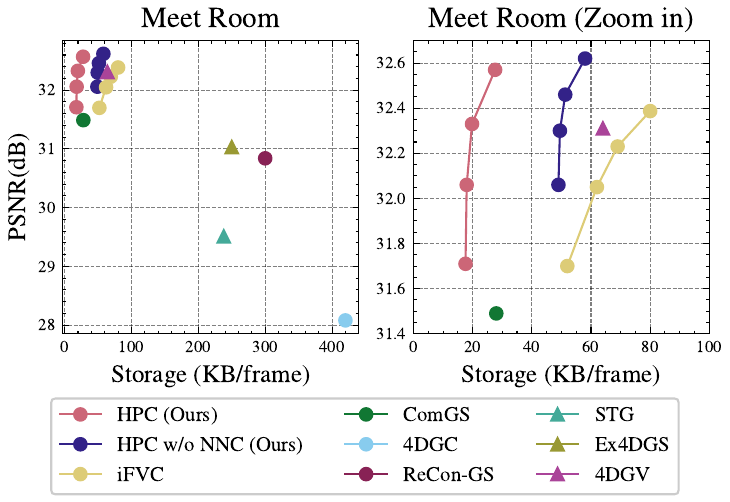}
	\caption{Rate-distortion curves on the N3DV and the Meet Room datasets. Methods marked with $\bigcirc$ and $\bigtriangleup$ respectively denote the online and offline optimized methods. Points and curves closer to the top-left corner indicate better performance.}
\label{fig:rdcurves}
\end{figure*}

\begin{figure*}[]
\centering
\includegraphics[width=0.245\linewidth]{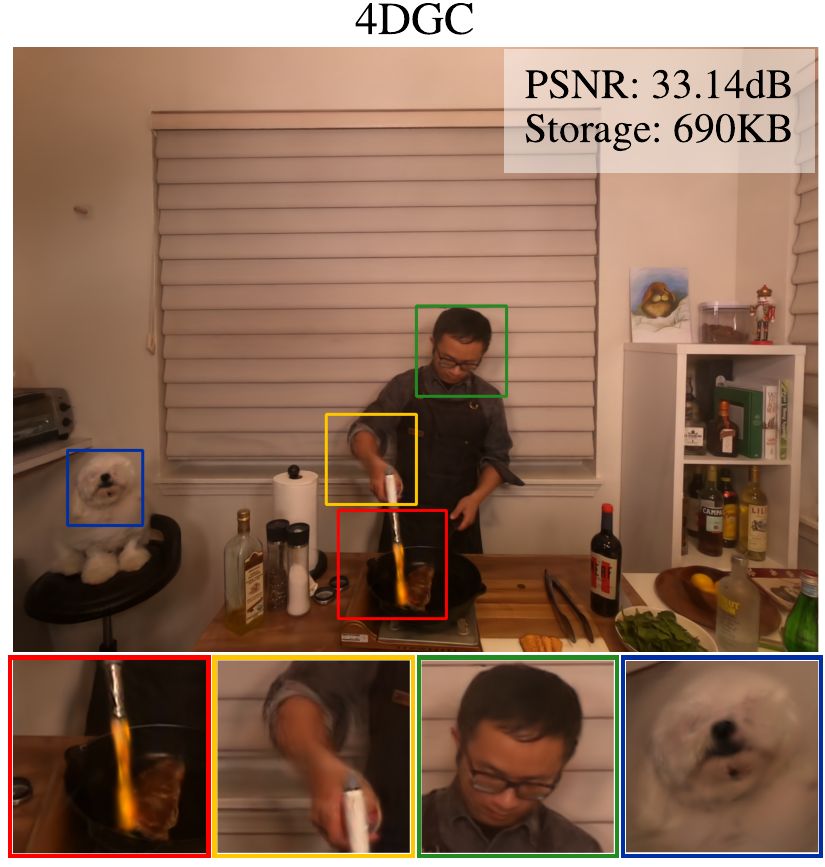}
\includegraphics[width=0.245\linewidth]{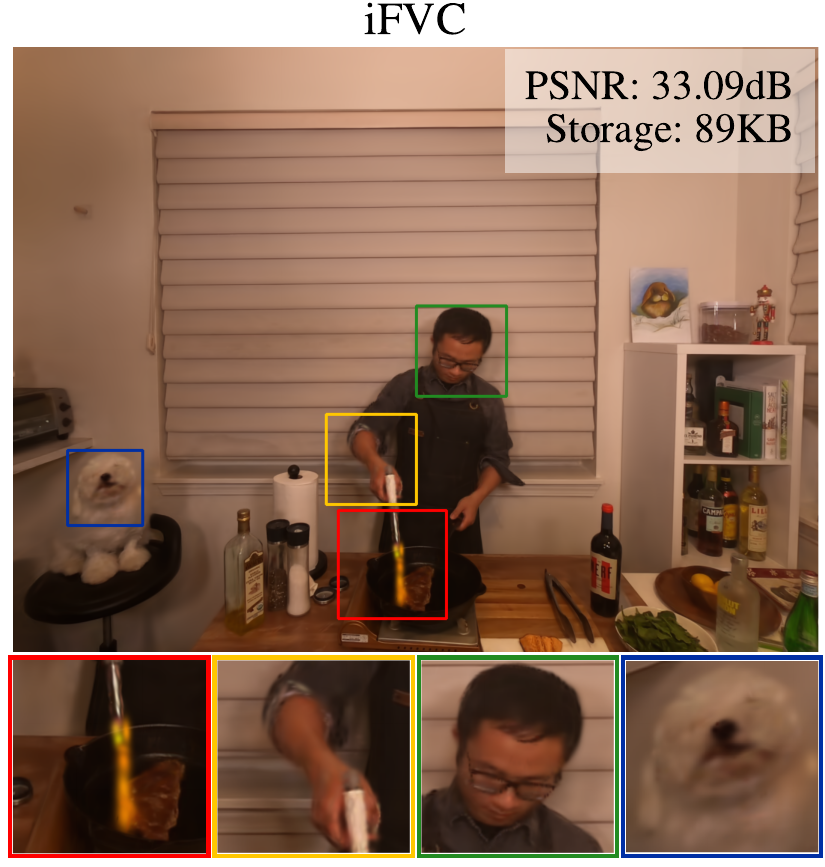}
\includegraphics[width=0.245\linewidth]{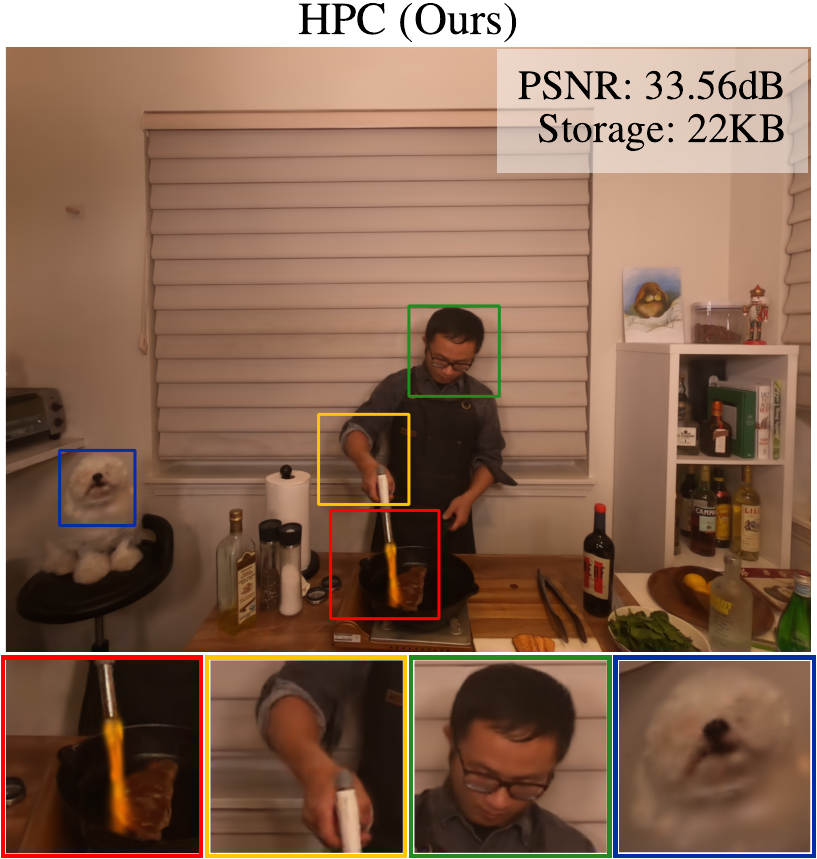}
\includegraphics[width=0.245\linewidth]{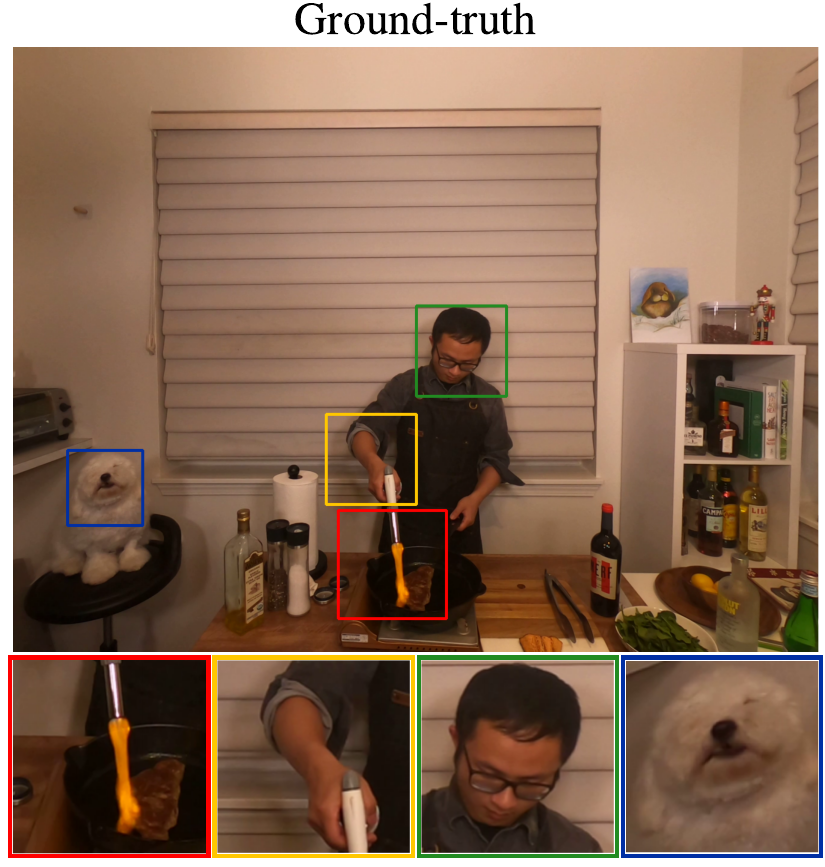}
\includegraphics[width=0.245\linewidth]{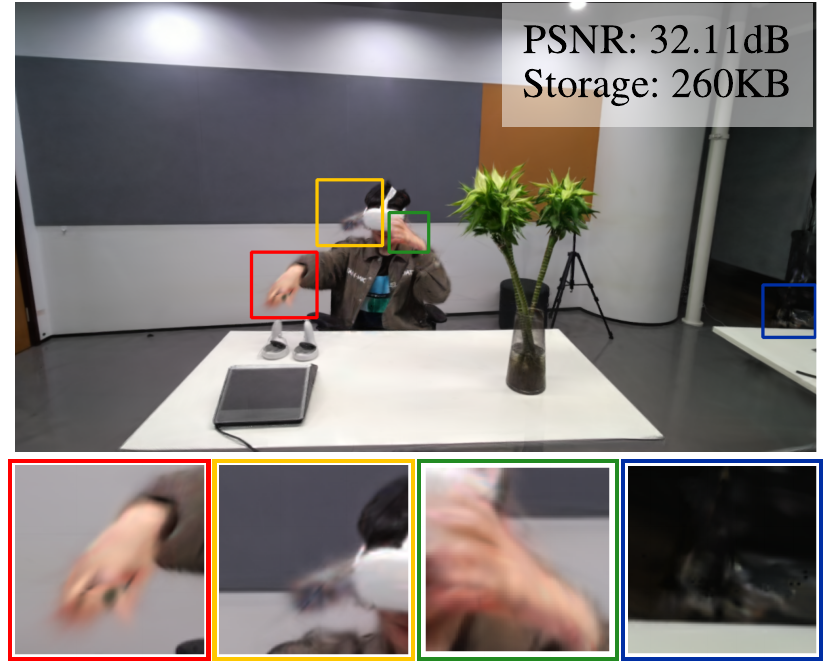}
\includegraphics[width=0.245\linewidth]{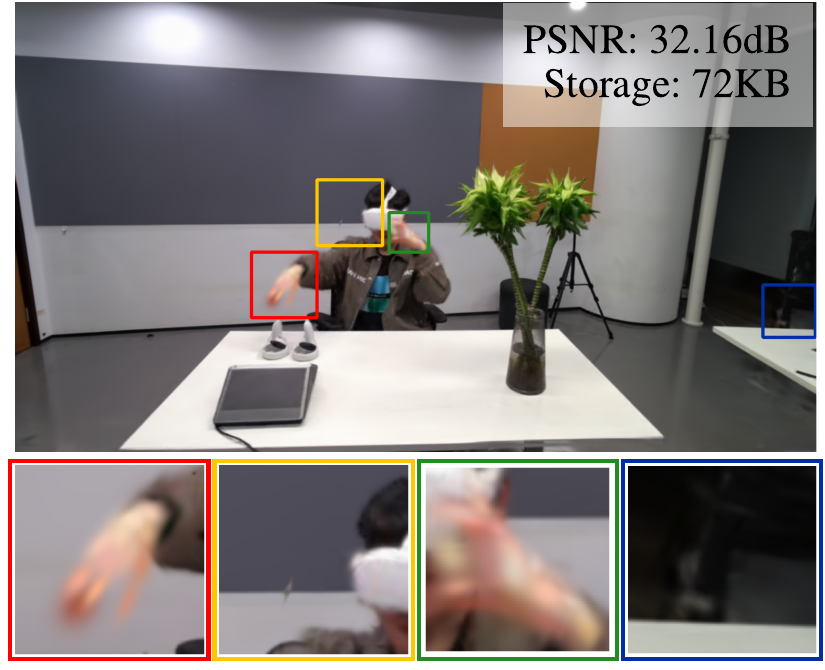}
\includegraphics[width=0.245\linewidth]{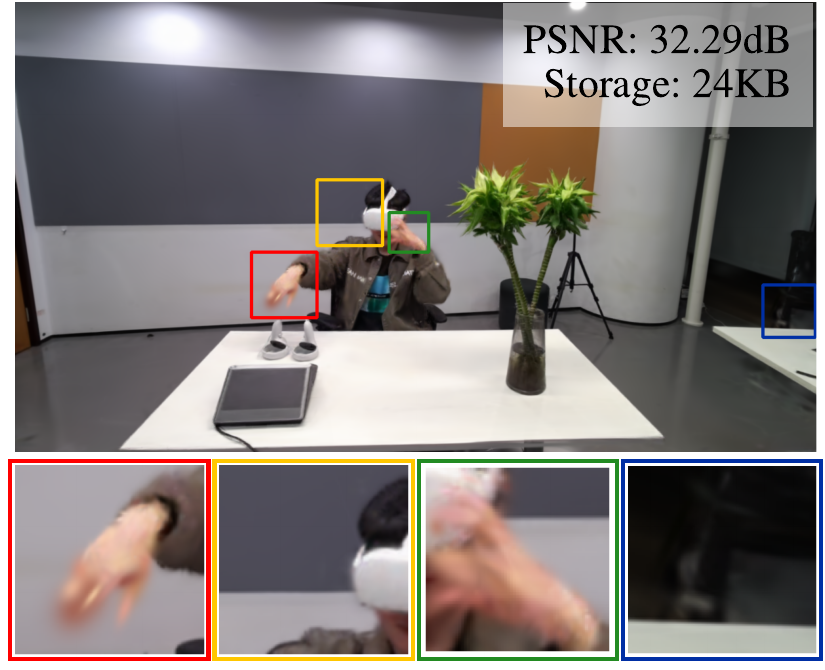}
\includegraphics[width=0.245\linewidth]{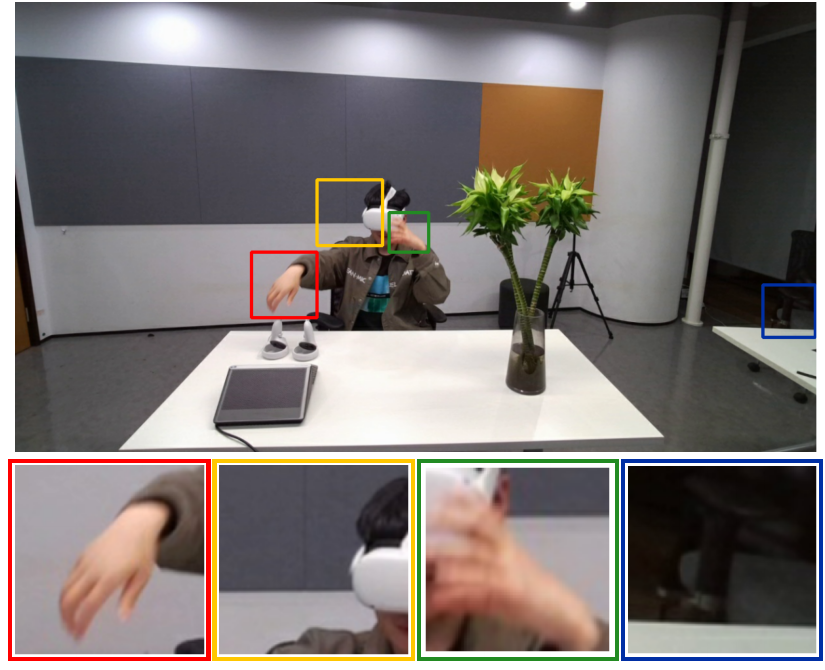}
\caption{Qualitative comparisons on the \textit{flame\_steak} and \textit{vrheadset}. For a fair comparison, we retrain all methods with the same initial sparse points, while iFVC~\cite{iFVC} and HPC share the same initial-frame model.}
\label{fig:quantitative}
\end{figure*}

\subsection{Experiment Setup}

\Paragraph{Datasets.}
We evaluate our method on two widely adopted FVV datasets: \textbf{(1) the N3DV dataset}~\cite{li2022neural}, captured by a 21-camera rig, which comprises dynamic scenes at 2704$\times$2028 and 30 FPS. 
\textbf{(2) the Meet Room dataset}~\cite{li2022streaming}, captured by a 13-camera system, comprising dynamic scenes at 1280$\times$720 and 30 FPS. For each dataset, one camera view is held out for testing, and the rest are used for training.

\Paragraph{Implementation Details.}
For the latent point hierarchy, we adopt the approach of ContextGS~\cite{contextgs} to construct $L=3$ scales, with the ratio of points between adjacent scales set to $0.2$.
\textcolor{black}{
The corresponding grid size $\epsilon$ is adaptively determined by binary search to approximately satisfy the predefined inter-scale point ratio.
}
The latent embeddings contain $16$ channels per scale. In the ILA module, we set $k=4$ for the $k$-NN search. For neural network compression, parameters are quantized to $B=8$ bit depth. The GOP size $T_{\text{GOP}}$ is set to $5$. 
For training, we use $\lambda\in\{0.048,0.032,0.016,0.002\}$ to obtain four rate-distortion operating points.
We train HAC~\cite{hac} as the initial frame for 15000 iterations following iFVC~\cite{iFVC}.
For the subsequent frames, we train our model for 500 iterations.
\textcolor{black}{
We use Adam with $\beta_1=0.9$, $\beta_2=0.999$, and no weight decay, with the learning rate decreasing from $5\times10^{-3}$ to $4.28\times10^{-3}$ within each frame.
}
We employ the torchac library~\cite{mentzer2019practical} for arithmetic coding. 
All experiments are conducted on an NVIDIA GeForce RTX 3090 GPU.
\textcolor{black}{
The implementation code and trained initial-frame models will be released to facilitate reproducibility.
}

\Paragraph{Baselines.}
We evaluate the proposed HPC by comparing it with state-of-the-art online frameworks, including 3DGStream~\cite{sun20243dgstream}, HiCoM~\cite{hicom2024}, iFVC~\cite{iFVC}, 4DGC~\cite{4dgc}, QUEEN~\cite{girish2024queen}, ComGS~\cite{chen2025motion}, 4DGCPro~\cite{zheng2025dgcpro}, ReCon-GS~\cite{fu2025recongs}, \textcolor{black}{D-FCGS~\cite{zhang2026dfcgs}, and StreamSTGS~\cite{ke2026streamstgs}}.
Several offline methods~\cite{li2024spacetime,li2025gifstream,lee2024ex4dgs,zhang2024mega,dai20254dgv}, which are optimized within multiple frames, are also included in the benchmark to serve as reference points for a more comprehensive performance assessment.
As existing methods lack neural network compression, we introduce a corresponding baseline, HPC w/o NNC, by disabling this component in our full model. This allows it to be fairly compared against prior methods while isolating the contribution of our other components.
\textcolor{black}{
For a strictly controlled comparison, we retrain iFVC~\cite{iFVC} using the same initialization and training-view settings as HPC. The results of the other compared methods are taken from their original papers or provided by the corresponding authors.
}

\Paragraph{Metrics.}
We evaluate rate-distortion performance using Peak Signal-to-Noise Ratio (PSNR) and Structural Similarity Index Measure (SSIM)~\cite{ssim} as objective distortion metrics, with bitrate measured in kilobytes per frame (KB/frame). 
\textcolor{black}{
Throughout this paper, we use the decimal convention, i.e., 1 KB = 1000 bytes.
The reported storage of HPC is averaged
over the complete sequence, including the initial frame.
}
For a precise comparison of rate-distortion efficiency, we compute the Bjontegaard Delta Bit-Rate (BD-BR)~\cite{bjontegaard2001calculation}, which quantifies the bitrate savings at equivalent PSNR quality.
\textcolor{black}{
BD-BR is calculated only over the overlapping PSNR interval of the two RD curves, without extrapolation.
}
Note that a negative BD-BR value indicates decreased storage at the same fidelity compared to the anchor method, which is desirable.

\subsection{Comparison Results}

\Paragraph{Quantitative Comparisons.}
The \textcolor{black}{quantitative} comparison is elaborated in Table~\ref{tab:benchmark}.
HPC achieves competitive reconstruction quality while requiring about 20 KB/frame at minimum for storage on both datasets, which is significantly lower than other online-optimized methods.
By adjusting $\lambda$ during training, HPC is able to achieve the best fidelity on the Meet Room dataset and the second-best on the N3DV dataset among the online methods, while consuming less than 40 KB/frame for storage.
Notably, the reconstruction fidelity is significantly influenced by the initial frame, which is discussed in the supplementary material.
\textcolor{black}{
At a playback rate of 30 FPS, the evaluated HPC operating points correspond to transmission bitrates of approximately 4.32-9.36 Mbps.
}
Comprehensively, Fig.~\ref{fig:rdcurves} shows the overall rate-distortion performance, where we can observe our HPC significantly surpasses previous state-of-the-art methods.
For a fair comparison, we use iFVC [7], which shares the same initial frame with ours, as the anchor for the overall BD-BR comparison. 
\textcolor{black}{Even without neural network compression, HPC achieves an average BD-BR reduction of 28.0\% over iFVC, demonstrating the effectiveness of our representation and aggregation design. With neural network compression enabled, the full HPC further improves the average bitrate saving to 67.8\%.}

We give a comprehensive analysis of the significant bitrate reduction.
In contrast to methods like HiCoM~\cite{hicom2024}, ComGS~\cite{chen2025motion}, and ReCon-GS~\cite{fu2025recongs}, which directly optimize Gaussian attribute residuals, HPC adopts a different strategy: it optimizes a latent representation paired with a lightweight network to predict residuals.
This implicit formulation allows entropy constraints to be applied properly, leading to high compactness and compression efficiency.
Compared to approaches with the implicit dynamic model like iFVC~\cite{iFVC}, 4DGC~\cite{4dgc}, 4DGCPro~\cite{zheng2025dgcpro}, and QUEEN~\cite{girish2024queen}, HPC develops a more compact and efficient latent representation. When combined with the neural network compression, this leads to greater overall performance gains.

\Paragraph{Qualitative Comparisons.}
We conduct a qualitative comparison with 4DGC~\cite{4dgc} and iFVC~\cite{iFVC} on the \textit{flame\_steak} sequence and the \textit{vrheadset} sequence. To ensure a fair comparison, all methods are initialized with the same sparse point cloud. Additionally, iFVC~\cite{iFVC} and our HPC share an identical first-frame model.
As shown in Fig.~\ref{fig:quantitative}, HPC achieves the best reconstruction quality while attaining the smallest model size. Notably, HPC better preserves fine details, such as facial features, animal fur, and flame textures. Furthermore, it demonstrates superior capability in modeling components with large motions, including fast-moving arms and hands.
These results indicate that HPC accurately captures dynamic scene elements, maintains high-fidelity details in complex objects, and achieves a highly compact representation.

\subsection{Ablation Study}
\label{sec:abl}
To assess the impact of individual components, we conduct comprehensive ablation studies.
Specifically, our analysis is structured around the three main contributions: the latent representation, the latent aggregation scheme, and the neural network compression scheme.

\begin{table}[t]
\color{black}
\centering
\caption{Ablation results with the full HPC model. Positive values indicate higher bitrate than HPC.}
\label{tab:ablation_hpc_ref}
\setlength{\tabcolsep}{7pt}
\begin{tabular}{lccc}
\toprule
Method & N3DV & MeetRoom & Average \\
\midrule
HPC w/o NNC & +87.3\%  & +175.7\% & +131.5\% \\
HPC w/o ILA & +19.6\%  & +25.0\%  & +22.3\% \\
HPC w/o CLA & +24.3\%  & +28.3\%  & +26.3\% \\
\bottomrule
\end{tabular}
\end{table}

\begin{table}[]
\centering
\caption{Ablation study of different latent representations on the N3DV dataset. Tri-plane is the anchor for BD-BR.}
\begin{tabular}{@{}lccc@{}}
\toprule
 & Tri-plane & Hash-grid & \begin{tabular}[c]{@{}c@{}}Hierarchical points (Ours)\end{tabular} \\ \midrule
BD-BR (\%)$\downarrow$ & 0.0 & -6.7 & \textbf{-21.8} \\ \bottomrule
\end{tabular}
\label{tab:abllatent}
\end{table}

\begin{table}[t]
\centering  
\caption{Study on different bit depths of neural network quantization. The results are based on the Meet Room dataset with the highest bitrate.}
\begin{tabular}{@{}c|cccc@{}}
\toprule
Bit depth & \begin{tabular}[c]{@{}c@{}}PSNR\\ (dB)$\uparrow$\end{tabular} & \begin{tabular}[c]{@{}c@{}}Latent\\ Storage (KB)$\downarrow$\end{tabular} & \begin{tabular}[c]{@{}c@{}}Network\\ Storage (KB)$\downarrow$\end{tabular} & \begin{tabular}[c]{@{}c@{}}Total\\ Storage (KB)$\downarrow$\end{tabular} \\ \midrule
32 & \textbf{32.62} & 16 & 43 & 59 \\
16 & 32.60 & 16 & 21 & 37 \\
8 & 32.57 & 16 & 10 & \textbf{26} \\
4 & 7.31 & 60 & 5 & 65 \\ \bottomrule
\end{tabular}
\label{tab:bitdepth}
\end{table}

\begin{table}[t]
\centering
\caption{Ablation study of neural network compression on the Meet Room dataset. $M_0$ is the anchor method for BD-BR calculation.}
\begin{tabular}{@{}c|ccc@{}}
\toprule
Method & $\textit{M}_0$ & $\textit{M}_1$ & $\textit{M}_2$ \\ \midrule
Quantization & \checkmark & \checkmark & \checkmark \\
Entropy constraint &  & \checkmark  & \checkmark \\
Temporal reference &  &  & \checkmark  \\ \midrule
BD-BR (\%)$\downarrow$ & 0.0 & -6.6 & \textbf{-14.6}  \\ \bottomrule
\end{tabular}
\label{tab:nnc}
\end{table}

\Paragraph{Different Types of Representation.}
We first compare the proposed hierarchical latent point with two established structural \textcolor{black}{representations}: the hash-grid and the tri-plane.
 The hash-grid adopts the same configuration as in 3DGStream~\cite{sun20243dgstream}, while the tri-plane adopts the same configuration as in iFVC~\cite{iFVC}.
As summarized in Table~\ref{tab:abllatent}, the proposed hierarchical points representation achieves the best performance among the three candidates.
This result demonstrates that our design achieves more efficient parameter allocation for discrete Gaussians compared to structural representations.

\Paragraph{Latent Aggregation.}
\begin{figure*}[]
	\centering
    \includegraphics[height=1.25in]{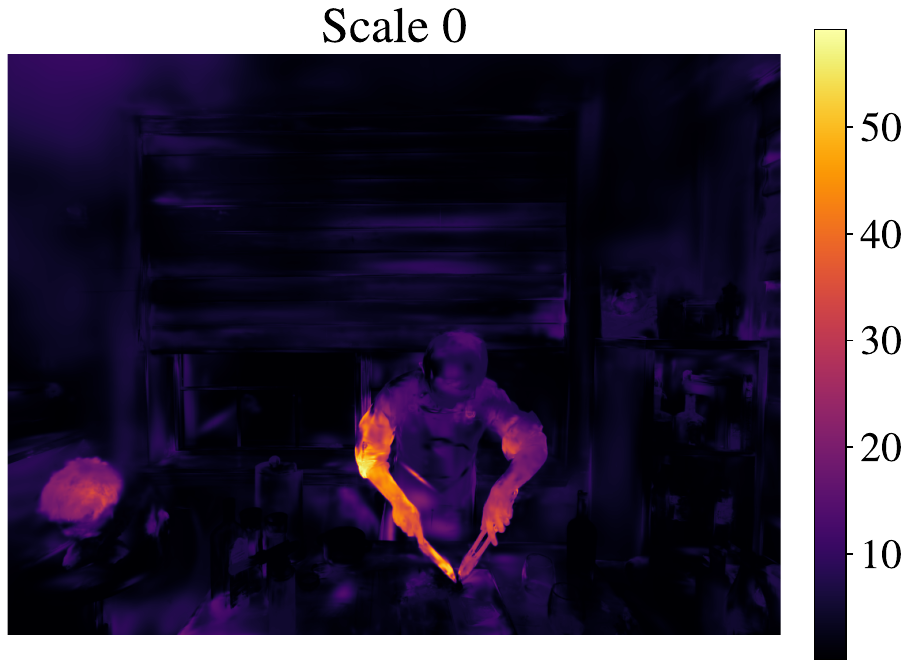}
    \includegraphics[height=1.25in]{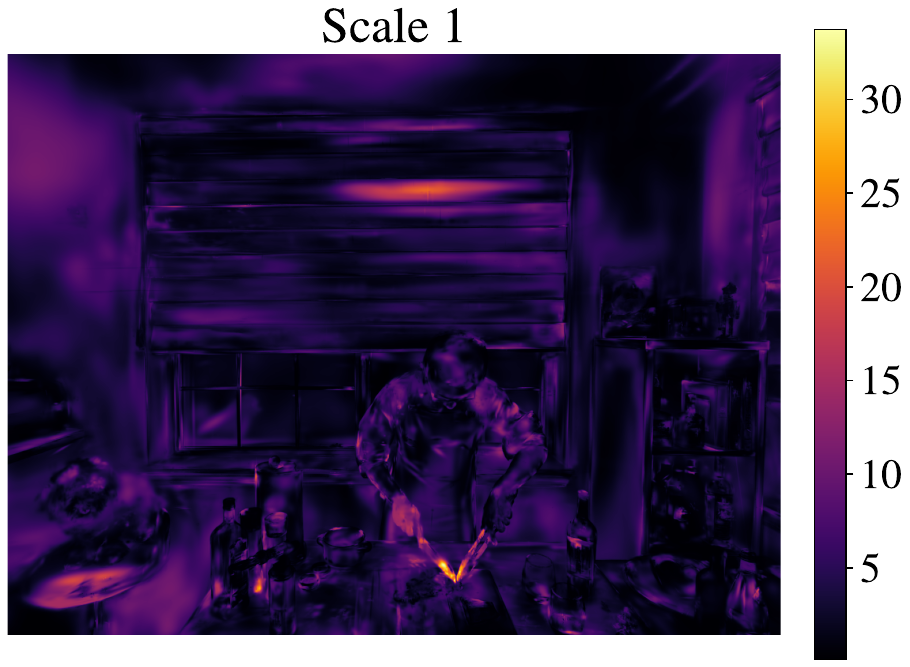}
    \includegraphics[height=1.25in]{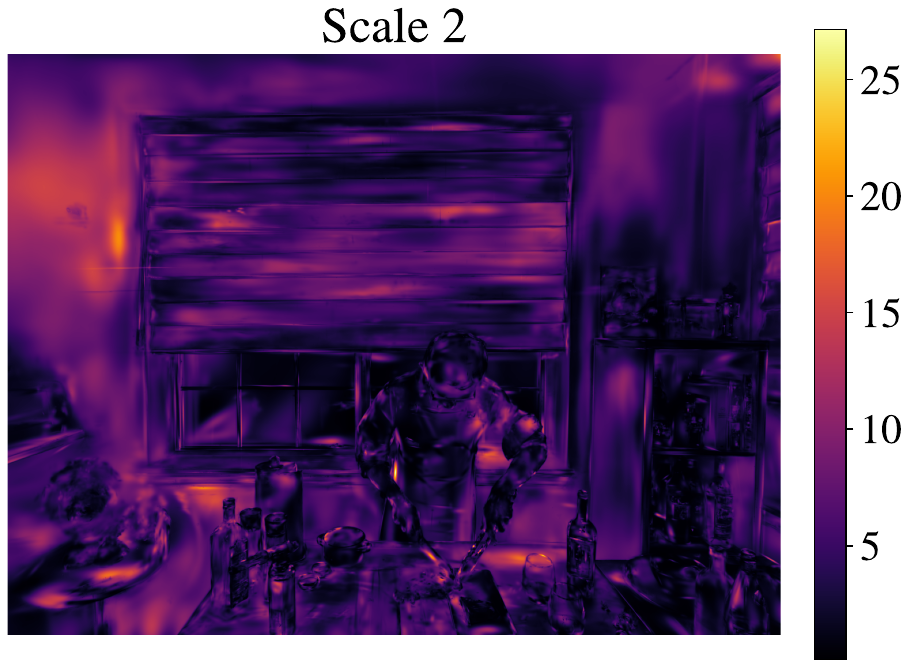}
    \includegraphics[height=1.25in]{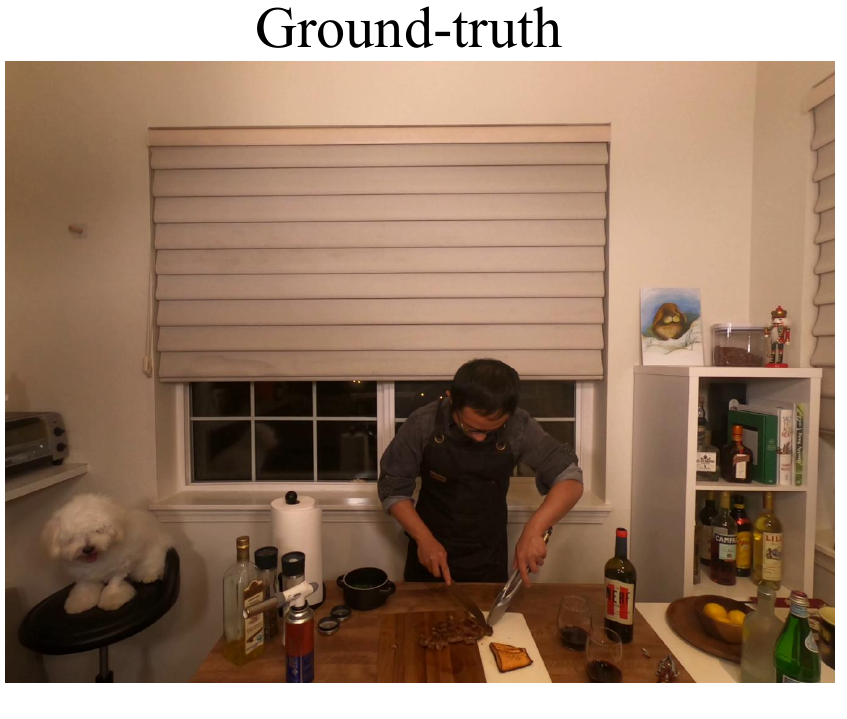}
	\caption{Bit allocation of latent embeddings in different scales. This visualization maps the bitrate consumed by each latent embedding to a color value, which is then rendered onto a 2D image to show the bit allocation.}
\label{fig:bpp_vis}
\end{figure*}
\textcolor{black}{
We evaluate the proposed aggregation scheme by separately ablating the ILA and CLA modules, as summarized in Table~\ref{tab:ablation_hpc_ref}. 
For the {w/o ILA} variant, the kNN-based intra-scale neighborhood aggregation is removed, and the original latent embeddings from different scales are directly concatenated without local neighborhood fusion. This results in BD-BR increases of 19.6\% and 25.0\% on N3DV and MeetRoom, respectively, corresponding to an average increase of 22.3\%. 
For the {w/o CLA} variant, the hierarchical representation is reduced to a single scale with 20 latent channels, such that the overall parameter count remains approximately comparable to that of the full model. This variant increases the BD-BR by 24.3\% on N3DV and 28.3\% on MeetRoom, with an average increase of 26.3\%. 
These results demonstrate the effectiveness of both local neighborhood aggregation and multi-scale feature aggregation in improving the rate-distortion efficiency.}
\textcolor{black}{
More extensive ablations on the number of scales, latent-channel dimension, neighborhood size, and inter-scale point ratio are provided in the supplementary material.}

\Paragraph{Neural Network Compression.}
\label{abl:nnc}
Prior to compression, the parameters must first be quantized. To investigate its effect, we reduce the precision from the original 32 bits to several lower bit depths, including 16, 8, and 4 bits. As shown in Table~\ref{tab:bitdepth}, the storage cost of network parameters decreases linearly with the quantization bit depth. Compared to the original precision, reconstructions at 8 or 16 bits exhibit only marginal degradation (less than 0.1 dB). However, at 4-bit quantization, the training becomes unstable and diverges, leading to a significant drop in reconstruction quality. Concurrently, the excessive quantization of the factorized entropy model causes the increasing bitrate of the latent representation. Based on this analysis, we select 8-bit quantization as our operating point, which effectively minimizes storage overhead while preserving the essential neural network capability. 

For further ablation study, we progressively incorporate other components into the baseline with 8-bit quantization, which is denoted as $\textit{M}_0$.
As shown in Table~\ref{tab:nnc}, to compress the parameters efficiently, we incorporate an entropy constraint into the parameter optimization, denoted as $\textit{M}_1$. 
This is implemented using the differentiable network-free Gaussian entropy model~\cite{zhang2024boosting} to enable bitrate estimation and entropy coding, which yields a \textcolor{black}{6.6\% BD-BR reduction relative to $\textit{M}_0$}.

Finally, we introduce temporal reference and residual coding across adjacent frames, denoted as $\textit{M}_2$. 
\textcolor{black}{
As shown in Table~\ref{tab:temporal_parameter_analysis}, after temporal-reference optimization, the reconstruction-related parameters exhibit clear element-wise correlation across consecutive frames, with an overall correlation coefficient of 0.78. More importantly, the empirical entropy decreases from 6.06 bits/parameter for the full parameters to 4.45 bits/parameter for the temporal residuals, corresponding to a 26.54\% reduction. These results quantitatively demonstrate that the resulting temporal residuals provide a more compact representation for entropy coding.
}

\textcolor{black}{However, recursive temporal prediction may accumulate errors
when the prediction chain becomes excessively long. We therefore
introduce the GOP reference structure with periodic parameter
refresh and investigate the impact of the GOP size.}
\textcolor{black}{
To further quantify the propagation of parameter quantization
errors, we compare the standard HPC with a full-precision
counterpart in which the quantization/dequantization of the
temporally predicted network parameters is disabled while
keeping the same temporal prediction and optimization procedure.
For the $k$-th position within a GOP, we define the
quantization-induced quality gap as
$\Delta Q(k)=\mathrm{PSNR}_{\mathrm{FP}}(k)
-\mathrm{PSNR}_{\mathrm{HPC}}(k)$.
As shown in Table~\ref{tab:gop_quant_error}, for the default GOP
size of 5, the gap increases from 0.067~dB at the starting
position to 0.125~dB at the last position, corresponding to
an additional degradation of only 0.058~dB across the GOP.
A similarly small increase of 0.078~dB is observed for GOP 10,
whereas the incremental gap rises substantially to 1.899~dB
for GOP 20.
}

With the GOP reference structure, \textcolor{black}{$\textit{M}_2$ achieves an overall BD-BR reduction of 14.6\% relative to the quantization-only $\textit{M}_0$ baseline}, as shown in Table~\ref{tab:nnc}.
\textcolor{black}{
Since motion intensity may vary across frames, a content-adaptive GOP strategy could further improve this trade-off by inserting I-frames according to the magnitude of the temporal parameter residuals. We leave the design of such an adaptive policy for future work.
}
\textcolor{black}{
Additional per-frame results for all evaluated scenes and a quantitative analysis of quantization-error propagation under different GOP sizes are provided in the supplementary material.
}

\subsection{Analysis}
\label{sec:analysis}

\begin{table}[t]
\color{black}
\centering
\caption{\textcolor{black}{Temporal correlation and empirical entropy of the temporally coded network parameters. Entropy is measured on quantized symbols in bits per parameter.}}
\label{tab:temporal_parameter_analysis}
\setlength{\tabcolsep}{4pt}
\begin{tabular}{lccc}
\toprule
Parameter Module
& Corr. $\uparrow$
& $H(\mathrm{Full})$
& $H(\mathrm{Residual})$
\\
\midrule
Feature/Offset MLP
& 0.79 & 5.91 & 3.94 \\
Anchor Pos./Rot. MLP
& 0.78 & 6.44 & 5.48 \\
ILA Position-Weight MLP
 & 0.81 & 5.43 & 3.94 \\
ILA Output MLP
 & 0.74 & 6.33 & 5.59 \\
\midrule
Overall
 & 0.78 & 6.06 & 4.45 \\
\bottomrule
\end{tabular}
\end{table}


\begin{table*}[t]
\color{black}
\centering
\caption{\textcolor{black}{Complexity comparison. The results are evaluated on an NVIDIA RTX 3090 GPU and averaged over the N3DV and MeetRoom datasets. Total encoding time measures the complete process from training to bitstream generation. The kNN construction time is included in the Train and Decode time. Rendering FPS is measured independently of decoding.}}
\label{tab:complexity}
\setlength{\tabcolsep}{3.5pt}
\begin{tabular}{lcccccc}
\toprule
Method
& Train
& kNN
& Bitstream Encode
& Total Encode
& Decode
& Render \\
& (s/frame)$\downarrow$
& (s/frame)$\downarrow$
& (s/frame)$\downarrow$
& (s/frame)$\downarrow$
& (s/frame)$\downarrow$
& (FPS)$\uparrow$ \\
\midrule
4DGC
& 49.81 & -- & 0.70 & 50.51 & 0.11 & 184 \\

iFVC
& 10.22 & -- & 0.10 & 10.32 & 0.09 & 140 \\

HPC w/o NNC
& 36.03 & 0.04 & 0.13 & 36.16 & 0.14 & 141 \\

HPC
& 60.61 & 0.04 & 0.33 & 60.94 & 0.22 & 140 \\
\bottomrule
\end{tabular}
\end{table*}


\begin{table}[t]
\color{black}
\centering
\caption{\textcolor{black}{Quantization-induced quality gap at different GOP sizes. The incremental gap measures the additional gap from the GOP starting position to the last position.}}
\label{tab:gop_quant_error}
\setlength{\tabcolsep}{5pt}
\begin{tabular}{ccccc}
\toprule
GOP
& $\Delta_Q(0)$
& Last $\Delta_Q$
& Incremental Gap
& Linear Slope \\
& (dB) & (dB) & (dB) & (dB/position) \\
\midrule
5  & 0.067 & 0.125 & +0.058 & +0.012 \\
10 & 0.089 & 0.167 & +0.078 & +0.008 \\
20 & 0.207 & 2.106 & +1.899 & +0.103 \\
\bottomrule
\end{tabular}
\end{table}

\Paragraph{Complexity.}
\textcolor{black}{
Table~\ref{tab:complexity} compares the computational complexity of HPC with 4DGC~\cite{4dgc} and iFVC~\cite{iFVC}. We separately report the per-frame training time and the subsequent bitstream encoding time, where the latter includes quantization and entropy coding after optimization. Total encoding time measures the complete process from per-frame training to bitstream generation. HPC requires 60.94~s/frame for the complete encoding process, including 60.61~s/frame for training and 0.33~s/frame for bitstream encoding.
}
\textcolor{black}{
Therefore, HPC is intended for offline encoding followed by streaming transmission and playback.}
For training time, the extended duration of HPC is attributed to its rate-aware training scheme for network parameters, which introduces more complex gradient updates and thus slows the optimization.
When the neural network compression module is removed, the training speed of HPC surpasses 4DGC~\cite{4dgc} but remains slower than iFVC~\cite{iFVC}.
This is primarily due to the additional parameters in the introduced aggregation modules and entropy model, which in turn require more iterations for convergence.
\textcolor{black}{
The kNN construction takes 0.043 s/frame on average with a peak optimization memory of 3.77 GiB. For substantially larger scenes, approximate kNN or voxel-hash-based lookup could be adopted to improve scalability.
}
The rendering is slightly slower for both HPC and iFVC~\cite{iFVC} due to the overhead of Neural Gaussian Splatting~\cite{scaffoldgs}, whereas 4DGC~\cite{4dgc} uses the faster, non-neural original splatting~\cite{3dgs}.
The longer overall coding time of HPC stems from its need to entropy encode both latent embeddings and network parameters, whereas the other two methods compress only the former.
When focusing specifically on the shared task of encoding latent embeddings, HPC matches the coding speed of iFVC~\cite{iFVC} and outperforms 4DGC~\cite{4dgc}.

\begin{figure}[t]
	\centering
	\includegraphics[width=0.9\linewidth]{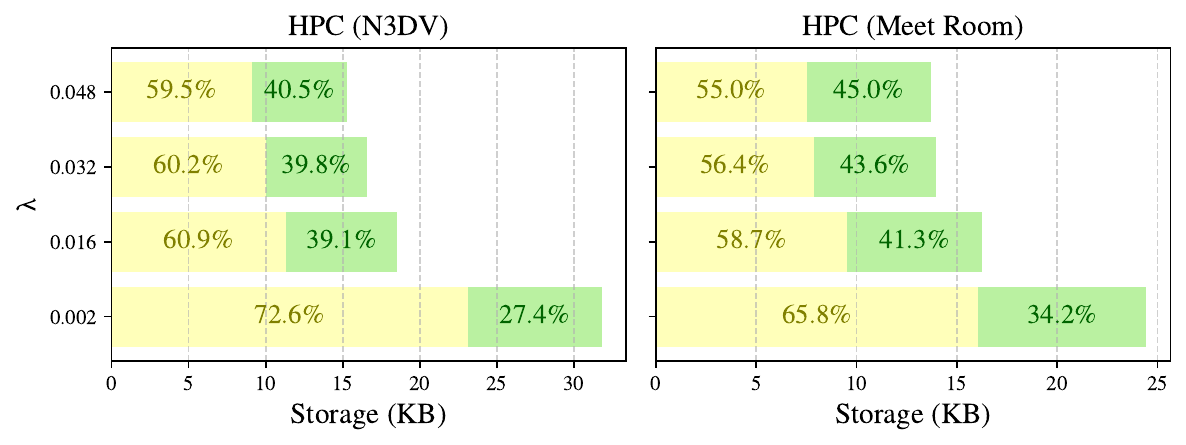}
    \includegraphics[width=0.9\linewidth]{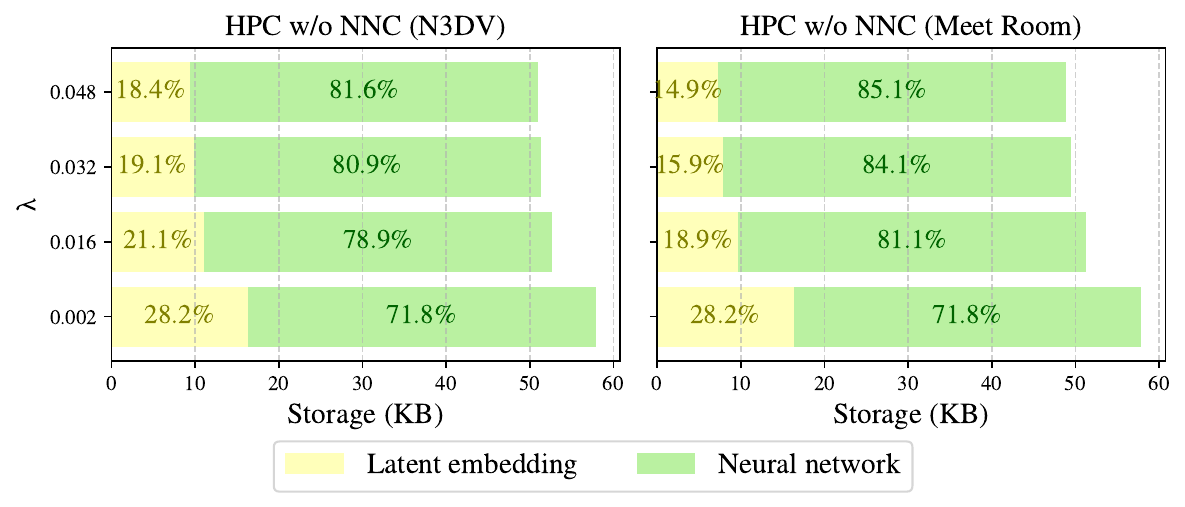}
	\caption{\textcolor{black}{Bit allocation between latent embedding and neural network. Metadata required for each component is included in the corresponding bitrate.}}
\label{fig:bit_alloc}
\end{figure}


\Paragraph{Bit Allocation.}
\textcolor{black}{The complete bitstream consists of the latent stream and the network-parameter stream, together with the metadata required for their decoding. The latent metadata include the quantization range, while the network metadata include the tensor-wise quantization statistics, Gaussian mean/standard deviation, and the layer-wise scaling factors $\eta$ for GOP P-frames. In the following analysis, these metadata bits are included in their corresponding latent or network components.
We provide a detailed breakdown of a representative GOP in the supplementary material.}

\textcolor{black}{
As shown in Fig.~\ref{fig:bit_alloc}, the bitrate allocation between the latent embeddings and network parameters varies across operating points. At lower bitrates, the two components consume comparable rates, whereas the proportion allocated to the latent embeddings increases with the total bitrate and reaches approximately 70\% at the highest bitrate. Without neural network compression, the network parameters account for around 80\% of the bitrate at low and medium rate points, demonstrating the importance of explicitly compressing the network parameters.
}





\section{Conclusion and future works}
\label{sec:conclusion}
We explore a hierarchical point-based latent representation tailored for dynamic Gaussian Splatting and introduce HPC. 
It ensures efficient parameter allocation, enables the effective capture of spatial correlations from non-uniformly distributed Gaussians, and ultimately delivers efficient, high-quality reconstruction.
A fully end-to-end compression framework, which jointly optimizes both the latent representation and the neural network for an optimal rate-distortion trade-off, is designed to achieve significant bitrate savings.
Extensive experiments demonstrate that HPC achieves the state-of-the-art compression performance against existing methods.
Comprehensive studies analyze and demonstrate the effectiveness of its technical components.
By achieving significant bitrate savings for dynamic Gaussian Splatting, our work paves the way for the field of streaming free-viewpoint video.

The primary limitation of HPC is its longer training duration and higher coding latency compared to competing methods due to the additional rate-aware optimization and compression for network parameters. 
\textcolor{black}{
In addition, the anchor set remains fixed after the initial frame, which limits the ability to explicitly represent completely new geometry outside the support of the existing anchors.
}
Future work should incorporate lightweight designs to accelerate both the training and coding processes, \textcolor{black}{and explore temporally adaptive
anchor growth and pruning with efficient signaling of anchor insertion
and removal}, thereby facilitating practical deployment.



\bibliographystyle{IEEEtran}
\bibliography{IEEEabrv,reflist.bib}

\vspace{-40pt}
\begin{IEEEbiography}[{\includegraphics[width=1in,height=1.25in,clip,keepaspectratio]{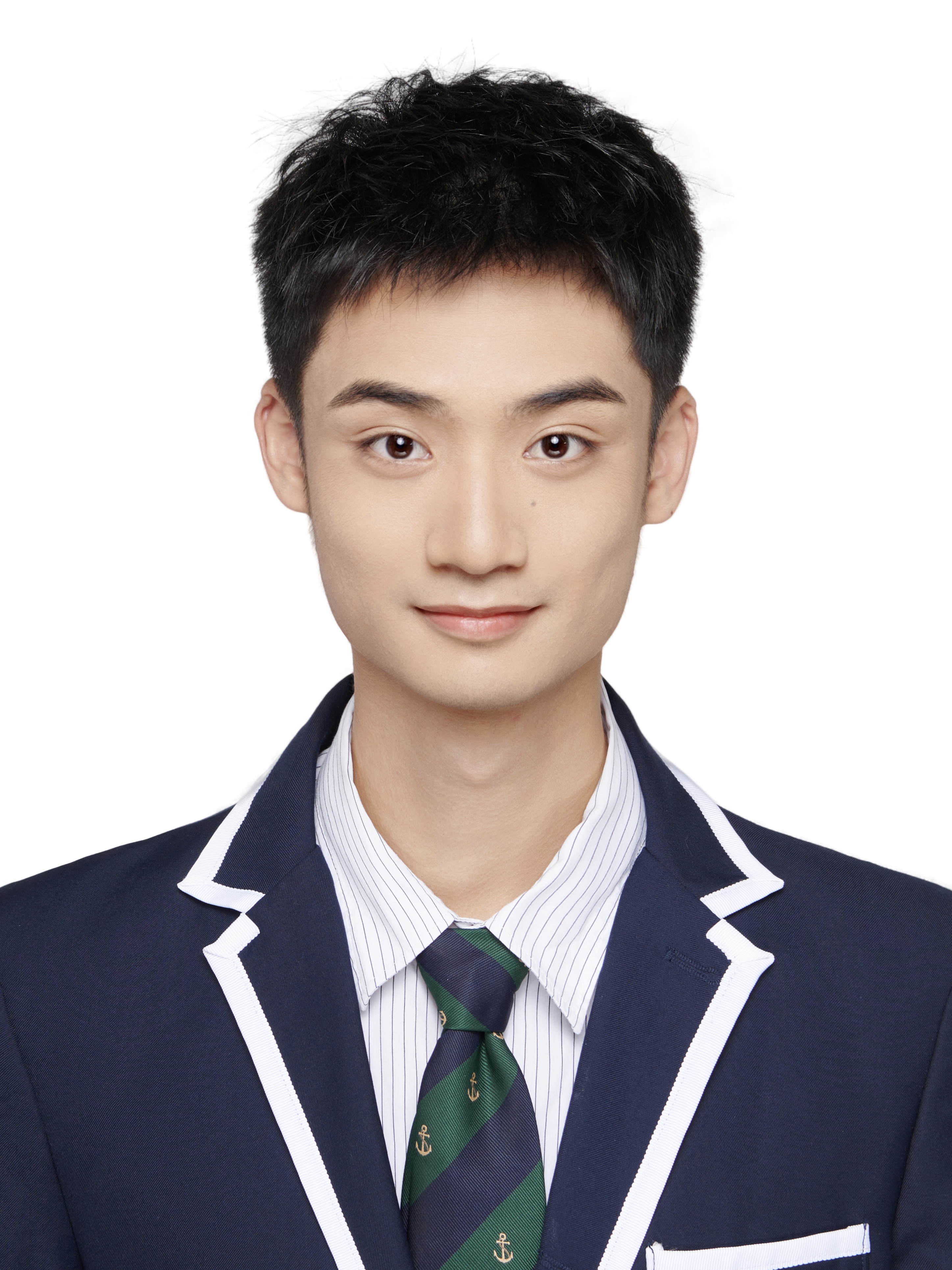}}]{Yangzhi Ma}
 received the B.S. degree from the Communication University of China, Beijing, China, in 2024. He is currently pursuing the M.S. degree in the University of Science and Technology of China, Hefei, China. His research interests include 3D data compression and processing.
\end{IEEEbiography}

\vspace{-25pt}
\begin{IEEEbiography}[{\includegraphics[width=1in,height=1.25in,clip,keepaspectratio]{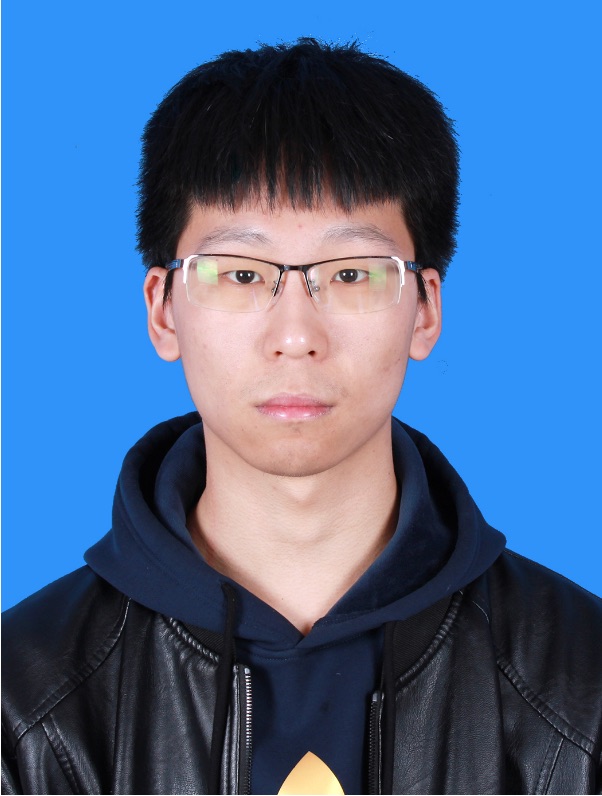}}]{Bojun Liu}
received the B.S. degree from Beijing University of Posts and Telecommunications, Beijing, China, in 2022. He is currently pursuing the Ph.D. degree in the University of Science and Technology of China, Hefei, China. His research interests include point cloud geometry compression and processing.
\end{IEEEbiography}

\vspace{-25pt}
\begin{IEEEbiography}[{\includegraphics[width=1in,height=1.25in,clip,keepaspectratio]{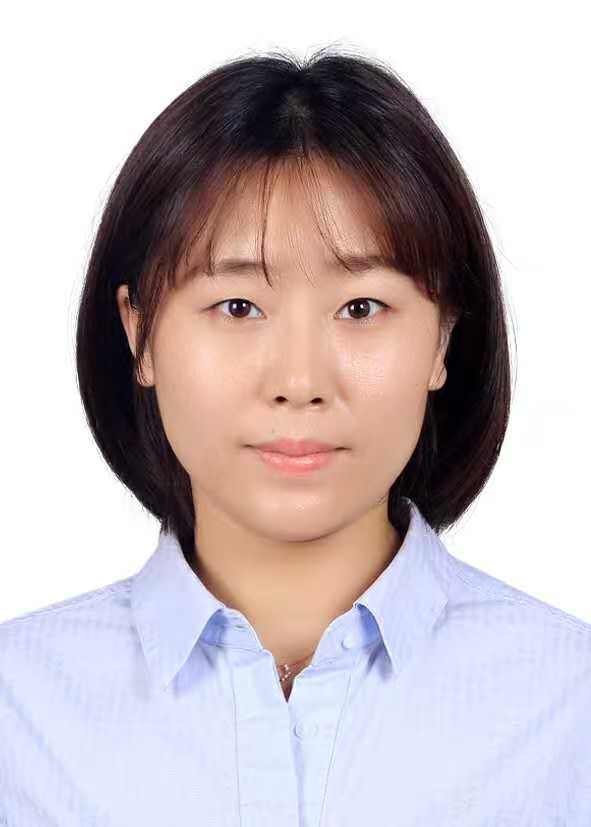}}]{Wenting Liao}
, a Ph.D. candidate in Electronic Engineering at the University of Science and Technology of China,  Product Director at the General Hospital of the People's Liberation Army. Her research interests include medical artificial intelligence, multimodal large models, and personalized recommendations.
\end{IEEEbiography}

\begin{IEEEbiography}[{\includegraphics[width=1in,height=1.25in,clip,keepaspectratio]{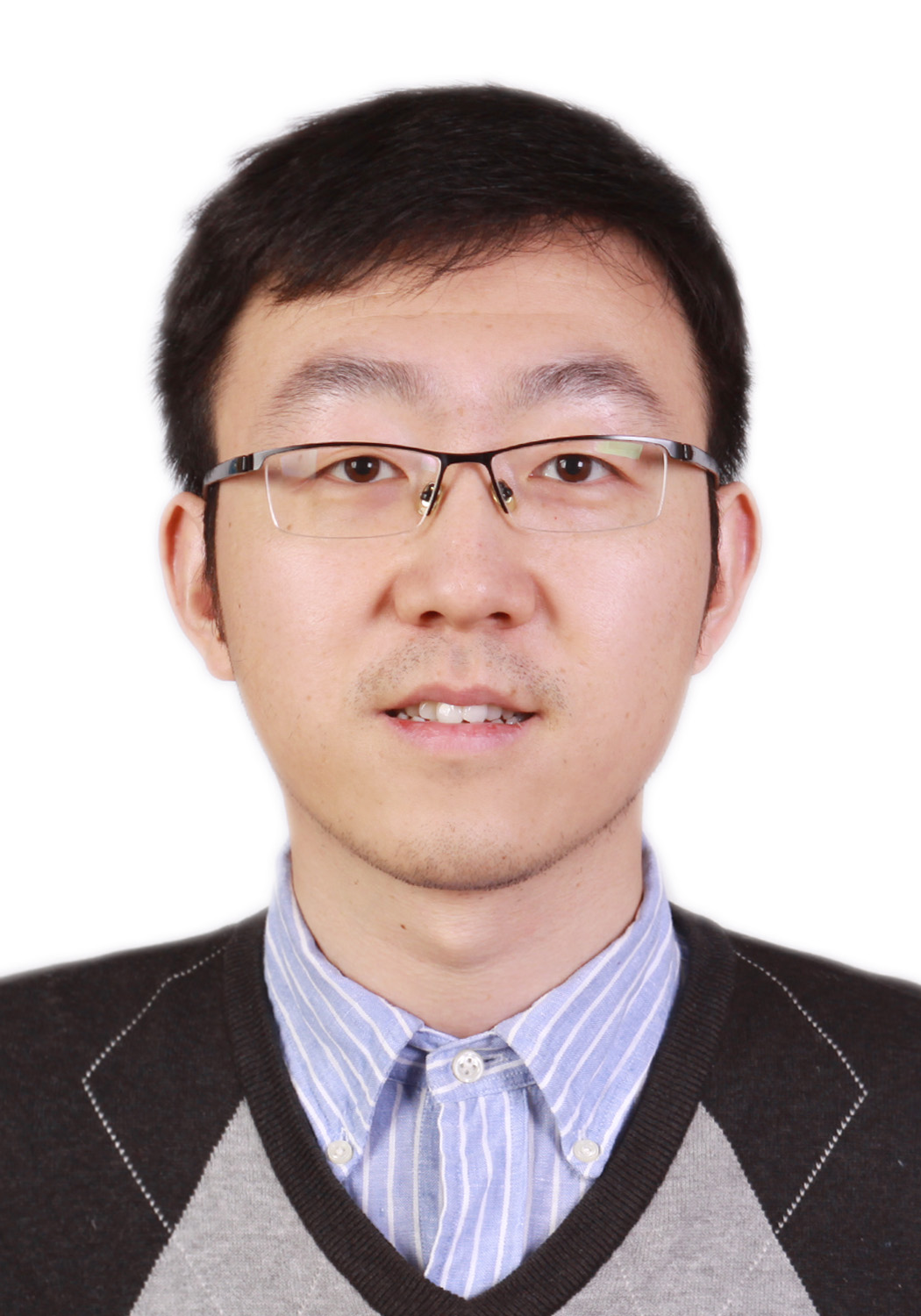}}]{Dong Liu}
(M'13--SM'19) received the B.S. and Ph.D. degrees in electrical engineering from the University of Science and Technology of China (USTC), Hefei, China, in 2004 and 2009, respectively. He was a Member of Research Staff with Nokia Research Center, Beijing, China, from 2009 to 2012. He joined USTC as a faculty member in 2012 and became a Professor in 2020.
His research interests include image and video processing, coding, analysis, and data mining.
He has authored or co-authored more than 200 papers in international journals and conferences. He has more than 30 granted patents. He has several technique proposals adopted by standardization groups. \par
He received the 2009 \textsc{IEEE Transactions on Circuits and Systems for Video Technology} Best Paper Award, VCIP 2016 Best 10\% Paper Award, and ISCAS 2022 Grand Challenge Top Creativity Paper Award. He and his students were winners of several technical challenges held in ISCAS 2023, ICCV 2019, ACM MM 2019, ACM MM 2018, ECCV 2018, CVPR 2018, and ICME 2016. He is a Senior Member of CCF and CSIG, and an elected member of MSA-TC of IEEE CAS Society. He serves or had served as the Chair of IEEE 1857.11 Standard Working Subgroup (also known as Future Video Coding Study Group), an Associate Editor for \textsc{IEEE Transactions on Image Processing}, a Guest Editor for \textsc{IEEE Transactions on Circuits and Systems for Video Technology}, an Organizing Committee member for VCIP 2022, ChinaMM 2022, ICME 2021, etc.
\end{IEEEbiography}

\vspace{-25pt}
\begin{IEEEbiography}[{\includegraphics[width=1in,height=1.25in,clip,keepaspectratio]{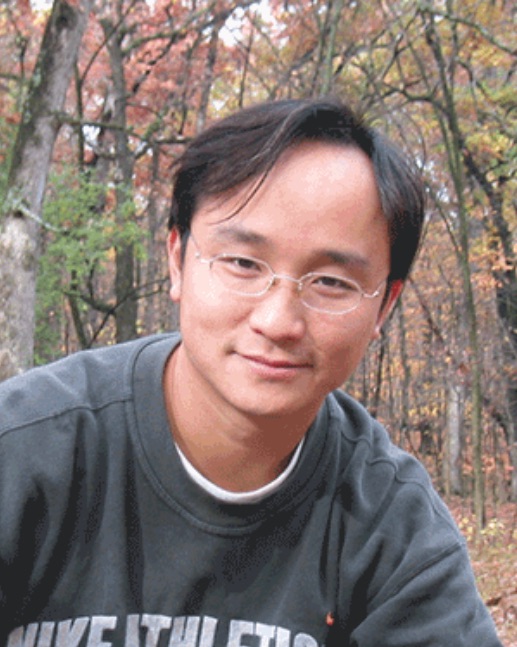}}]{Zhu Li}
is a full professor with the Dept of Computer Science \& Electrical Engineering, University of Missouri, Kansas City(UMKC), and the director of NSF I/UCRC Center for Big Learning (CBL) at UMKC. He received his PhD in Electrical \& Computer Engineering from Northwestern University in 2004. He was AFRL summer faculty at the UAV Research Center, US Air Force Academy (USAFA), 2016-18, 2020-23. He was Senior Staff Researcher with the Samsung Research America's Multimedia Standards Research Lab in Richardson, TX, 2012-2015, Senior Staff Researcher with FutureWei Technology's Media Lab in Bridgewater, NJ, 2010-2012, Assistant Professor with the Dept of Computing, the Hong Kong Polytechnic University from 2008 to 2010, and a Principal Staff Research Engineer with the Multimedia Research Lab (MRL), Motorola Labs, from 2000 to 2008. His research interests include point cloud and light field compression, graph signal processing and deep learning in the next gen visual compression, remote sensing, image processing and understanding. He has 50+ issued or pending patents, 200+ publications in book chapters, journals, and conferences in these areas. He is an IEEE senior member, Associate Editor-in-Chief for IEEE Trans on Circuits \& System for Video Tech, 2020, Associate Editor for IEEE Trans on Image Processing(2020), IEEE Trans.on Multimedia (2015-18), IEEE Trans on Circuits \& System for Video Technology(2016-19). He received the Best Paper Runner-up Award at the Perception Beyond Visual Spectrum (PBVS) grand challenge at CVPR 2023, Best Paper Award at IEEE Int'l Conf on Multimedia \& Expo (ICME), Toronto, 2006, and IEEE Int'l Conf on Image Processing (ICIP), San Antonio, 2007.
\end{IEEEbiography}

\vspace{-25pt}
\begin{IEEEbiography}[{\includegraphics[width=1in,height=1.25in,clip,keepaspectratio]{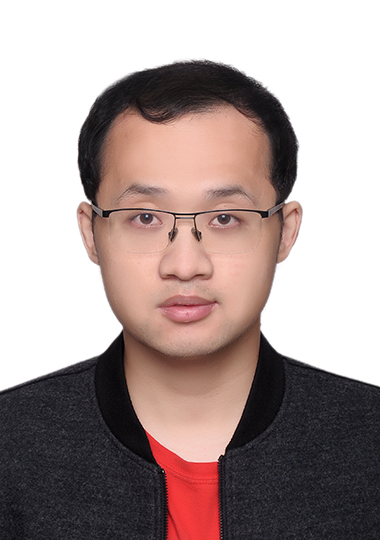}}] {Li Li} (M'17-SM'25) received the B.S. and Ph.D. degrees in electronic engineering from University of Science and Technology of China (USTC), Hefei, Anhui, China, in 2011 and 2016, respectively.
He was a visiting assistant professor in University of Missouri-Kansas City from 2016 to 2020.
He joined the department of electronic engineering and information science of USTC as a research fellow in 2020 and became a professor in 2022. \par
His research interests include image/video/point cloud coding and processing.
He has authored or co-authored more than 100 papers in international journals and conferences. 
He has more than 20 granted patents. 
He has several technique proposals adopted by standardization groups.
He received the Multimedia Rising Star 2023.
He received the Best 10\% Paper Award at the 2016 IEEE Visual Communications and Image Processing (VCIP) and the 2019 IEEE International Conference on Image Processing (ICIP).
He serves as a senior associate editor for \textsc{IEEE Transactions on Circuits and Systems for Video Technology} since 2026 and an associate editor for \textsc{IEEE Transactions on Multimedia} since 2025. 
\end{IEEEbiography}

\end{document}